\documentclass[10pt,twocolumn,letterpaper]{article}

\usepackage{wacv}              % To produce the CAMERA-READY version
\definecolor{wacvblue}{rgb}{0.21,0.49,0.74}
\usepackage{multicol}
\usepackage{multirow}
\usepackage{graphicx}
\usepackage{subcaption}
\usepackage{pifont}
\usepackage{amsmath}
\usepackage{microtype}
\usepackage[table]{xcolor}
\usepackage{pgfplots}
\pgfplotsset{compat=1.18}

\definecolor{goodgreen}{RGB}{0,150,40}
\definecolor{neutralgray}{RGB}{120,120,120}

\definecolor{wacvblue}{rgb}{0.21,0.49,0.74}
\usepackage[pagebackref,breaklinks,colorlinks,allcolors=wacvblue]{hyperref}

\def\wacvPaperID{678} % *** Enter the WACV Paper ID here
\def\confName{WACV}
\def\confYear{2027}

\title{UT-HMR: Uncertainty-Guided Temporal Modeling for Robust 3D Human Mesh Recovery}

\author{Zhibo Wang, Sirisha Rambhatla, Yuhao Chen, Maria Koshkina, David Clausi\\
{\tt\small \{zhibo.wang, sirisha.rambhatla, yuhao.chen1, mkoshkina, dclausi\}@uwaterloo.ca}\\
University of Waterloo, 
Waterloo, ON, Canada, N2L 3G1\\
}

\begin{document}
\maketitle

\begin{abstract}
Monocular 3D human mesh recovery (HMR) remains unreliable under weak
observations caused by long camera-to-person distances, occlusion, and motion
blur. In challenging scenarios such as sports, these factors make the
problem highly ambiguous, as different 3D poses may yield similar 2D
projections. This ambiguity is especially severe for one-stage HMR, which
recovers all meshes directly from the full image without person crops or
external 2D cues. To handle this ambiguity, we present UT-HMR (\textbf{U}ncertainty-guided
\textbf{T}emporal HMR), to our knowledge the first one-stage HMR framework to
capture correlated pose uncertainty across joints. Each person query predicts
a Gaussian distribution over the full SMPL pose with a compact
low-rank-plus-diagonal covariance, which is converted into joint-wise
precision weights for heteroscedastic supervision while keeping inference
deterministic. Temporal early fusion further injects neighboring-frame
evidence before pose decoding, supplying the visual evidence needed to resolve
the ambiguity that the uncertainty captures. With a lightweight ViT-B
backbone, UT-HMR achieves state-of-the-art results on VIP-HockeyPose-v2 (ice
hockey) and WorldPose (soccer), reducing MPJPE, PA-MPJPE, and PVE by up to
$4.7\%$, $3.8\%$, and $5.6\%$, and remains competitive on 3DPW against methods
with larger backbones and higher input resolutions.
\end{abstract}    
\section{Introduction}
\label{sec:intro}

\begin{figure}[t!]
    \centering
    \includegraphics[width=0.95\linewidth]{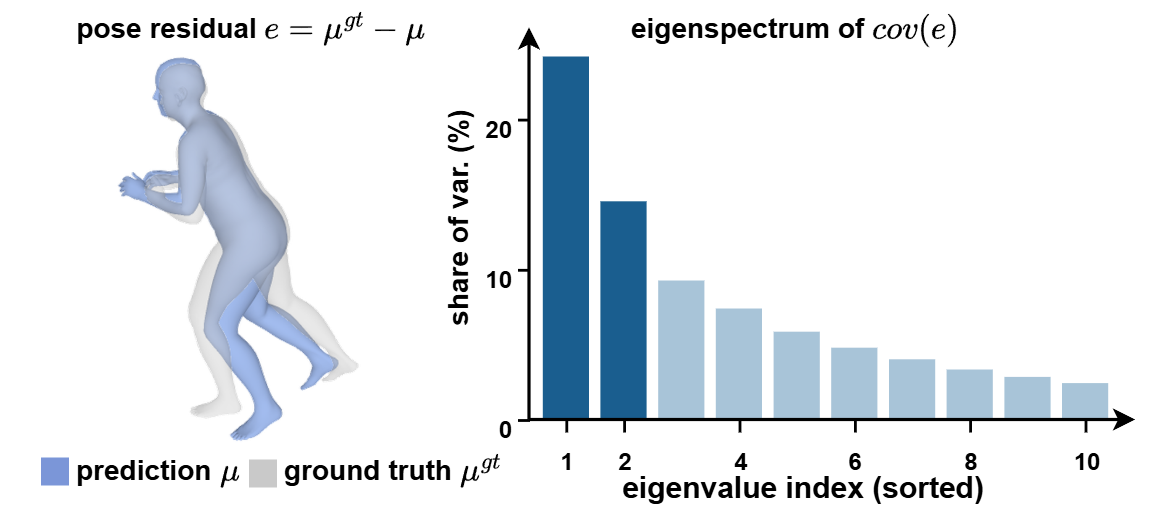}
    \caption{Empirical covariance spectrum of per-person 6D pose residuals ($24$ joints $\times\,6 = 144$
dimensions) on
    the ice hockey dataset. The top two directions explain 39\% of the total variance.}
    \label{fig:lowrank}
    \vspace{-3mm}
\end{figure}

\begin{figure*}[t]
  \centering
  % ---- 第一行:a b c 三张并列 ----
  \begin{subfigure}[t]{0.32\linewidth}
    \includegraphics[width=\linewidth]{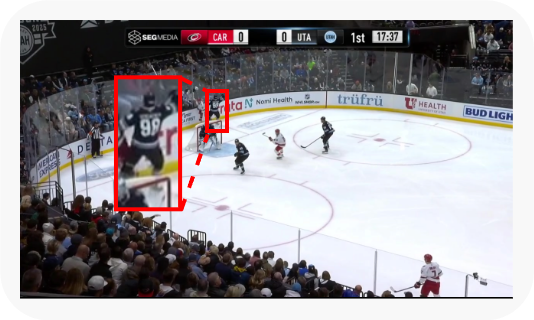}
    \caption{Input Frame}
    \label{fig:sub_a}
  \end{subfigure}
  \hfill
  \begin{subfigure}[t]{0.32\linewidth}
    \includegraphics[width=\linewidth]{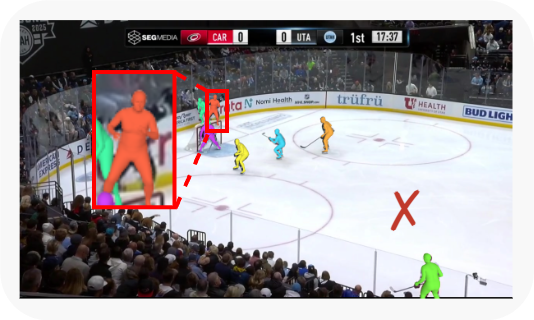}
    \caption{SAT-HMR \cite{SAT-HMR}}
    \label{fig:sub_b}
  \end{subfigure}
  \hfill
  \begin{subfigure}[t]{0.32\linewidth}
    \includegraphics[width=\linewidth]{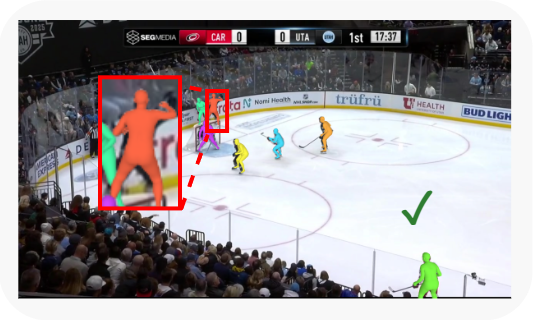}
    \caption{UT-HMR (Ours)}
    \label{fig:sub_c}
  \end{subfigure}

  \vspace{2mm} % 行间距，可调

  % ---- 第二行:d 一张，宽度等于上面三张之和 ----
  \begin{subfigure}[t]{\linewidth}
    \includegraphics[width=\linewidth]{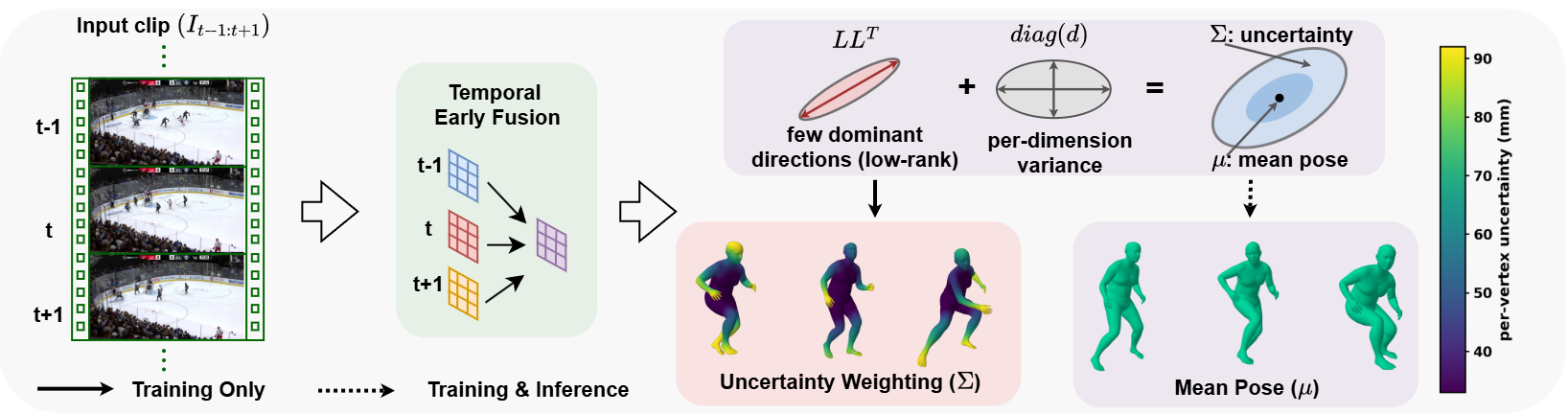}
\caption{
How UT-HMR handles projection ambiguity. The input clip ($T{=}3$) is encoded
through temporal early fusion. The pose mean $\mu$ is used for deterministic
mesh reconstruction, while the predicted covariance models pose uncertainty and
provides joint-wise weights for uncertainty-aware training.
}
    \label{fig:sub_d}
  \end{subfigure}

\caption{
\textbf{Resolving projection ambiguity in one-stage HMR.}
A distant or occluded player admits multiple plausible 3D poses from a single 2D
observation: (a) the target player (red box) is small and partially occluded;
(b) single-frame SAT-HMR~\cite{SAT-HMR} predicts an incorrect pose, while
(c) UT-HMR reconstructs it correctly. (d) Temporal early fusion gathers
multi-frame evidence, and the pose is modeled as a Gaussian: the mean $\mu$
yields the deterministic mesh, and the covariance $\Sigma$ provides joint-wise
uncertainty weights during training.
}

\label{fig:main}
\end{figure*}

3D human mesh recovery (HMR) from monocular images is inherently
ill-posed: the mapping from 2D observations to 3D pose is highly ambiguous,
making mesh quality strongly dependent on the visual evidence in the input
(Fig.~\ref{fig:main}). This ambiguity is particularly severe for one-stage HMR
methods~\cite{MultiHMR,SAT-HMR}, which recover multiple meshes directly from the
full image without cropped person regions used by two-stage methods
\cite{gvhmr,HMR2.0,WHAM} as additional evidence. Nevertheless, the one-stage
paradigm is particularly attractive for real-world multi-person applications:
it removes the need for person-level detection, jointly processes all people
from the full image, and supports efficient multi-person inference
~\cite{ROMP,MultiHMR,SAT-HMR}. These properties are especially valuable in crowded scenes like sports, where
the large number of players makes per-person cropping and inference costly,
while the single-pass, end-to-end design keeps the pipeline simple and
fast~\cite{ROMP,BEV,AiOS,MultiHMR}. We therefore seek to improve the robustness of
one-stage HMR under ambiguous observations while preserving these practical
advantages.

A natural remedy is to model pose uncertainty, which has proven
effective in crop-based HMR. Prior works capture uncertainty over probabilistic
2D pose features~\cite{r1-full}, pre-extracted 2D keypoint and depth
cues~\cite{r2-full}, or joint locations~\cite{r3-full}, or learn
multi-hypothesis distributions~\cite{ProHMR,MHPro}. None transfers directly to
one-stage HMR: they depend on crops, tracking, or explicit 2D inputs, and
hypothesis-based inference is ill-suited to estimating all people jointly.

% \textcolor{blue}{We bridge this gap with a design tailored to the one-stage setting:
% each person query predicts its pose distribution without requiring crops or
% auxiliary cues, while the covariance guides training without introducing
% additional inference hypotheses. Our statistical analysis (Fig.~\ref{fig:lowrank})
% shows that the baseline pose error concentrates largely in a few dominant
% directions. Accordingly, we model the predicted pose as a Gaussian distribution
% whose covariance explicitly represents full-body pose uncertainty, using a
% compact low-rank-plus-diagonal parameterization to capture correlated and
% residual uncertainty. The covariance provides joint-wise precision weights
% during training, while inference remains deterministic. Since uncertainty can
% measure but not reduce ambiguity, we further introduce temporal early fusion
% to inject neighboring-frame evidence before pose decoding. Unlike crop-based
% temporal aggregation~\cite{r4-full,r5-full}, our fusion operates on full-image
% features within the one-stage pipeline, providing additional evidence to resolve
% the ambiguity reflected by the predicted uncertainty.}

We bridge this gap with a design tailored to the one-stage setting.
Our statistical analysis (Fig.~\ref{fig:lowrank}) shows that the pose error of
a one-stage baseline concentrates largely in a few dominant directions.
Accordingly, we model each person's pose as a Gaussian whose compact
low-rank-plus-diagonal covariance captures correlated and residual uncertainty.
This distribution is predicted directly from each person query of the one-stage
decoder, requiring no crops or auxiliary cues; the covariance provides
joint-wise precision weights during training, while inference remains
deterministic. Since uncertainty can measure but not reduce ambiguity, we
further introduce temporal early fusion, which injects neighboring-frame
evidence before pose decoding; unlike crop-based temporal
aggregation~\cite{TCMR,r5-full}, it operates on full-image features within the
one-stage pipeline.

% This dependence has been
% largely overlooked, as most prior work is evaluated in everyday
% settings~\cite{3DPW} or on synthetic data~\cite{AGORA}, where the target person is
% clearly visible. The situation changes in more demanding scenarios such as sports,
% e.g., ice hockey and soccer, where the large camera-to-player distance and the fast
% pace of play frequently degrade the observation through occlusion and motion blur;
% under such weak evidence, we find that current models become unreliable. In ice hockey, for instance, bulky equipment and blurred appearance make a player's orientation difficult to determine. As a result, a model may predict a 3D pose that differs markedly from the ground truth while producing nearly the same 2D projection, revealing a projection ambiguity that existing methods have largely overlooked. This ambiguity is especially severe for one-stage methods~\cite{MultiHMR,SAT-HMR},
% which recover meshes directly from full images without auxiliary cues such as
% tracking results or 2D keypoints that could help reduce pose ambiguity. An
% example is shown in Fig.~\ref{fig:main}.

We name our method UT-HMR (\textbf{U}ncertainty-guided \textbf{T}emporal HMR),
which combines temporal early fusion with full-pose uncertainty modeling
(Fig.~\ref{fig:main}); to our knowledge, UT-HMR is the first one-stage HMR
framework to capture correlated pose uncertainty across joints. We evaluate it
on ice hockey~\cite{SSE} and soccer~\cite{WorldPose}, where ambiguous
observations are common, as well as on the general-purpose 3DPW
benchmark~\cite{3DPW}. UT-HMR achieves state-of-the-art performance on both
sports benchmarks, reducing MPJPE, PA-MPJPE and PVE by up to $4.7\%$, $3.8\%$
and $5.6\%$ over the strongest baseline, while remaining competitive on 3DPW
against larger models with higher
input resolutions.

% We name our method UT-HMR (\textbf{U}ncertainty-guided \textbf{T}emporal HMR)
% and evaluate it on ice hockey and soccer, where ambiguous observations occur
% frequently, as well as on the general-purpose 3DPW
% benchmark~\cite{3DPW}, where the gains show that our approach also
% generalizes to everyday scenarios. 

% In summary, our contributions are
% threefold:
% \begin{itemize}
%      \item We propose a temporal early-fusion mechanism that injects
%     neighboring-frame evidence before pose decoding, reducing the potential ambiguity of single frame.
%     \item We introduce an uncertainty modeling in
%     one-stage HMR: a low-rank-plus-diagonal Gaussian covariance around the
%     deterministic pose estimate, motivated by the empirical structure of pose
%     uncertainty, whose precision weights drive heteroscedastic 3D joint
%     supervision.
%     \item UT-HMR achieves state-of-the-art performance on challenging ice
%     hockey and soccer benchmarks, while remaining competitive on
%     general-purpose 3DPW.
% \end{itemize}

\section{Related Work}
\label{sec:related_work}

\subsection{Multi-Person Human Mesh Recovery}
\label{subsec:multiperson_hmr}

Human mesh recovery (HMR) in multi-person scenarios, typically built on
parametric human models such as SMPL~\cite{SMPL} and SMPL-X~\cite{SMPL-X}, is
essential for real-world applications like sports
analysis~\cite{SSE, gen4d, avatar4d}. Existing approaches generally follow either a two-stage or a one-stage paradigm. Two-stage methods, such
as HMR~\cite{HMR} and CameraHMR \cite{CameraHMR}, first detect and crop each person and
then perform mesh regression. Although cropping preserves image
resolution and usually improves accuracy, the overall performance depends
heavily on the quality of the detector, making these methods less suitable
for crowded scenes. In contrast, one-stage methods jointly estimate all
people from the full image. ROMP~\cite{ROMP} and BEV~\cite{BEV} regress
pixel-aligned body-center and mesh-parameter maps using CNN backbones, while
recent transformer-based approaches formulate multi-person HMR as a set
prediction problem~\cite{DETR,DeformableDETR, AiOS}. Multi-HMR~\cite{MultiHMR}
employs a cross-attention human prediction head with one query per detected
person, AiOS~\cite{AiOS} progressively filters and expands queries, and
SAT-HMR~\cite{SAT-HMR} introduces scale-adaptive tokens for real-time
inference. Our method builds upon this one-stage paradigm and further
enhances the decoder queries with temporal context and uncertainty
modeling, producing more robust results under weak observations.

\subsection{Video-Based Human Mesh Recovery}
\label{subsec:temporal_hmr}

To exploit temporal information and improve motion consistency, video-based
methods aggregate features across multiple frames. VIBE~\cite{VIBE} employs
recurrent encoders to capture temporal dynamics, while
TCMR~\cite{TCMR} aggregates past, future, and whole-sequence features.
Attention-based methods such as MPS-Net~\cite{MPS-Net} and
GLoT~\cite{GLoT} further model long-range temporal dependencies, and
PSVT~\cite{PSVT} extends one-stage multi-person estimation to video with
progressive spatio-temporal transformers. Another line of work focuses on
recovering world-grounded motion. WHAM~\cite{WHAM} and TRAM~\cite{TRAM}
combine pose estimation with camera motion to reconstruct trajectories in the
world frame, while GVHMR~\cite{gvhmr} performs sequence modeling in a
gravity-aligned coordinate system.

These methods mainly target long-range motion modeling and temporal
consistency over entire sequences. In contrast, our goal is to alleviate the
projection ambiguity caused by weak single-frame observations rather than to
perform full video-based HMR. Therefore, we adopt a lightweight temporal
early-fusion strategy that incorporates information from neighboring frames
before pose decoding, supplying additional evidence while preserving the
efficiency of a one-stage multi-person framework.

\subsection{Probabilistic Human Mesh Recovery}
\label{subsec:prob_hmr}

Monocular human mesh recovery is inherently ill-posed due to depth ambiguity,
occlusion, and truncation. To characterize such ambiguity, probabilistic methods
predict distributions or multiple plausible hypotheses instead of deterministic
point estimates. ProHMR~\cite{ProHMR} models pose uncertainty using conditional
normalizing flows, while MH-HMR~\cite{MH-HMR} and MHPro~\cite{MHPro} generate
multiple hypotheses and refine them through inter- and intra-hypothesis
communication. More recently, HuMR~\cite{HuMR} utilizes uncertainty encoded in
2D pose heatmaps to supervise the learned 3D mesh distribution. Different from
these methods, which explicitly sample, rank, or aggregate multiple hypotheses,
UT-HMR keeps inference deterministic. We parameterize a lightweight low-rank-plus-diagonal
Gaussian distribution around the deterministic pose prediction and use the
predicted covariance to derive joint-wise precision weights for heteroscedastic
training, allowing uncertainty to guide training while preserving efficient
one-stage multi-person inference.
\section{Method}

\subsection{Preliminaries}
\label{sec:prelim}

In this work, we adopt SMPL~\cite{SMPL} to represent the 3D human body. Given an input image $\mathbf{I}\in\mathbb{R}^{H\times W\times 3}$, the SMPL model parameterizes a human body using a pose parameter $\theta\in\mathbb{R}^{24\times3}$ and a shape parameter $\beta\in\mathbb{R}^{10}$, and maps them to a 3D mesh:
\begin{equation}
\mathbf{V}=\mathrm{SMPL}(\theta,\beta), \qquad
\mathbf{V}\in\mathbb{R}^{6890\times3}.
\end{equation}

In practice, recent HMR methods~\cite{HMR2.0,gvhmr} employ a 6D rotation representation for body pose due to its superior continuity and optimization properties~\cite{144D}. Therefore, instead of directly regressing the axis-angle pose parameter $\theta$, the network predicts a pose representation: $\mu\in\mathbb{R}^{24\times6}$,
where each joint rotation is encoded by a 6D vector. The predicted rotations are subsequently converted to the axis-angle representation required by the SMPL model, yielding $\theta\in\mathbb{R}^{24\times3}$. Together with the shape parameter $\beta\in\mathbb{R}^{10}$, the recovered pose parameters are used to reconstruct the mesh $\mathbf{V}$:

\begin{equation}
\mathbf{V}
=
\mathrm{SMPL}(\Phi(\mu),\beta),
\end{equation}
where $\mu\in\mathbb{R}^{24\times6}$ denotes the predicted 6D pose representation, $\beta\in\mathbb{R}^{10}$ denotes the shape parameter, and $\Phi(\cdot)$ converts the 6D rotation representation into the axis-angle representation required by SMPL.

The 3D joint locations are then obtained from the mesh using a predefined linear regressor $\mathcal{R}\in\mathbb{R}^{J\times6890}$:
\begin{equation}
\mathbf{J}=\mathcal{R}\mathbf{V}, \qquad
\mathbf{J}\in\mathbb{R}^{J\times3},
\end{equation}
where $J$ denotes the number of target joints.

\subsection{Overall Pipeline}

Following recent one-stage HMR methods~\cite{MultiHMR, SAT-HMR}, we build our
one-stage multi-person framework upon a DETR-style architecture~\cite{DETR}. A ViT-B/14 encoder first extracts patch features from the input image and
produces a memory feature map
$\mathcal{M}\in\mathbb{R}^{N_m\times c}$, where $N_m$ and $c$ denote the
number of patch tokens and the feature dimension, respectively. A DETR-style
decoder with $N_q$ learnable queries then performs cross-attention over
$\mathcal{M}$. Each query predicts a confidence score, and queries whose
confidence exceeds a threshold (0.3 in our experiments) are treated as
individual persons. Lightweight prediction heads further regress the pose
parameter $\mu$ and shape parameter $\beta$, which are converted by the
SMPL model into a mesh with 6890 vertices and the corresponding 3D joints.

\begin{figure*}
    \centering
    \includegraphics[width=\linewidth]{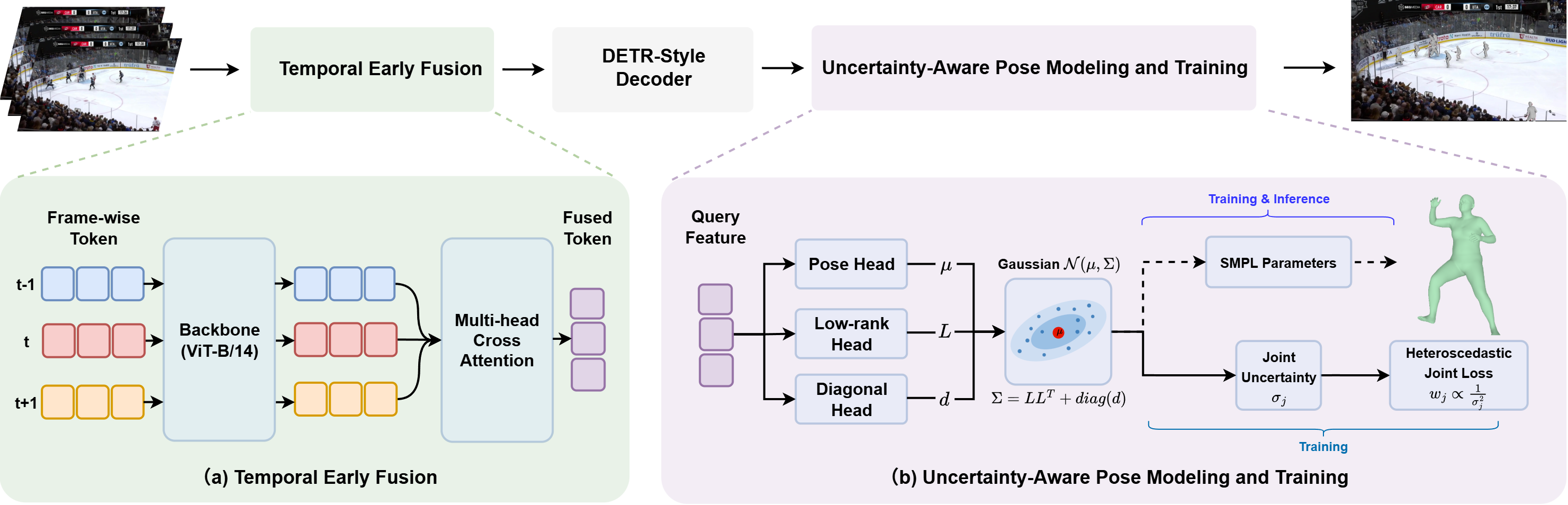}
\caption{
\textbf{Overall pipeline of UT-HMR.}
UT-HMR first fuses a short input clip through temporal early fusion and obtains per-person query features with a DETR-style decoder. The pose head predicts the deterministic pose mean $\mu$ for final mesh inference, while the low-rank and diagonal heads estimate a Gaussian covariance $\Sigma$ for uncertainty-aware training. The predicted covariance is supervised by the distribution loss and converted into joint-wise precision weights for the heteroscedastic joint loss.
}
\label{fig:overall_pipeline}
\end{figure*}

When directly applying this backbone to challenging sports scenarios such as
ice hockey, we observe a recurring failure mode. Under weak observations
caused by small scale, occlusion, and motion blur, the predicted pose often
produces a plausible 2D reprojection while exhibiting an incorrect 3D
configuration (Fig.~\ref{fig:main}). We attribute this phenomenon to projection ambiguity: a degraded 2D observation may correspond to
multiple plausible 3D poses, yet a deterministic single-frame model is forced
to commit to one of them. To address this issue, we introduce two
complementary components. First, a temporal early-fusion mechanism
(Sec.~\ref{sec:fusion}) injects short-term motion information before pose
decoding, providing additional evidence to alleviate ambiguity. Second, an uncertainty modeling branch (Sec.~\ref{sec:prob}) estimates the uncertainty associated with each prediction using a low-rank-plus-diagonal Gaussian covariance, which is further used to guide heteroscedastic training. Together, these two components enable more robust mesh recovery
under ambiguous observations. The overall pipeline is illustrated in
Fig.~\ref{fig:overall_pipeline}.

\subsection{Temporal Early Fusion}
\label{sec:fusion}

Challenging scenarios such as ice hockey and soccer often suffer from weak
per-frame observations, including small person scale, frequent occlusion, and
motion blur. Under such conditions, a single frame may not provide sufficient
evidence to resolve the projection ambiguity inherent to monocular HMR. Existing video-based HMR methods commonly exploit temporal information through temporal motion modeling, consistency constraints, or trajectory refinement \cite{WHAM,TRAM,TCMR,VIBE}. However, these strategies aggregate per-person features or refine predicted trajectories after cropping and tracking. As a result, they cannot directly provide full-image temporal evidence for an ambiguous center-frame observation within a one-stage pipeline.

We instead aggregate neighboring frames as auxiliary visual evidence before the
SMPL prediction heads. Given a short clip centered at frame $t$, we use
$T{=}3$, i.e., $\{I_{t-1}, I_t, I_{t+1}\}$. For each frame, the image encoder produces a memory feature map
$\mathcal{M}_{t}$. The three frame-wise memories are then concatenated into a
unified multi-frame memory for the center frame:
\begin{equation}
\tilde{\mathcal{M}}_t
=
\left[
\mathcal{M}_{t-1};
\mathcal{M}_{t};
\mathcal{M}_{t+1}
\right].
\label{eq:fusion}
\end{equation}
The decoder queries associated with the center frame cross-attend to
$\tilde{\mathcal{M}}_t$. In this way, each person query can selectively
aggregate complementary evidence from neighboring frames at the feature level,
rather than combining the outputs of separately predicted meshes.

Only the center frame is supervised during training, and the clip window slides
over the sequence at inference time to obtain frame-wise predictions. This design
introduces temporal evidence without requiring an explicit smoothness loss,
motion model, or hand-crafted fusion weights. Because fusion is performed before
pose regression, the enriched query features can improve the deterministic pose
estimate and provide stronger context for the uncertainty modeling branch in
Sec.~\ref{sec:prob}, where the remaining ambiguity is modeled by a low-rank-plus-diagonal
Gaussian covariance.

\subsection{Pose Uncertainty Modeling}
\label{sec:prob}

Although temporal fusion provides additional evidence to alleviate projection
ambiguity, the recovered pose can still remain uncertain under weak observations
such as occlusion, motion blur, and small person scale. Instead of treating all
predicted joints as equally reliable, we explicitly estimate the pose uncertainty
associated with each joint and use it to guide training.

Let $h\in\mathbb{R}^{768}$ denote the decoder feature of a matched query. The
deterministic pose head $f_{\theta}$ predicts a 6D pose
$\mu=f_{\theta}(h)\in\mathbb{R}^{144}$, corresponding to 24 joints with 6D
rotations. In parallel, an uncertainty branch models the pose as a Gaussian
centered at $\mu$ and predicts its covariance $\Sigma$. A full covariance over
$D=144$ dimensions would require $D(D+1)/2=10{,}440$ independent parameters per
query, which is unnecessarily large for monocular HMR. Human poses are highly
structured and can be effectively regularized by compact latent pose
priors~\cite{SMPL-X}. Our empirical analysis (Sec. \ref{sec:abla_hyper}) further shows that the learned
uncertainty remains concentrated in only a few dominant directions, suggesting
that pose uncertainty is intrinsically compact, while excessive covariance
dimensions may introduce noisy estimates. We therefore adopt a
low-rank-plus-diagonal covariance, where $L$ captures shared uncertainty factors
and $d$ models residual per-dimension uncertainty. This parameterization requires
only $Dr+D$ parameters ($432$ for $r=2$), over $20\times$ fewer than a full
covariance.

Specifically, a low-rank head predicts
$L\in\mathbb{R}^{144\times r}$ from the query feature $h$, modeling correlated
uncertainty that is shared across pose dimensions, while a diagonal head
predicts $d\in\mathbb{R}^{144}$, representing independent per-dimension
variance. The covariance is then assembled from the two heads as:
\begin{equation}
\Sigma = LL^\top+\operatorname{diag}(d),
\label{eq:gauss}
\end{equation}
which, together with the pose mean $\mu$, defines the predicted pose
distribution $\mathcal{N}(\mu,\Sigma)$ over the 6D pose space. The rank $r$
controls the number of correlated uncertainty directions modeled by the
low-rank term $LL^\top$; based on the empirical analysis above, we set $r=2$
in all our experiments.

% Specifically, a low-rank head predicts
% $L\in\mathbb{R}^{144\times r}$, which models correlated pose uncertainty across
% pose dimensions, while a diagonal head predicts $d\in\mathbb{R}^{144}$, which
% represents independent per-dimension variance. The covariance and pose
% distribution are then defined as
% \begin{equation}
% \begin{aligned}
% \Sigma
% &= LL^\top+\operatorname{diag}(d),\\
% Dist(h)
% &= \mathcal{N}(\cdot \mid \mu,\Sigma).
% \end{aligned}
% \label{eq:gauss}
% \end{equation}
% Here, $Dist(h)$ denotes the predicted pose distribution, and $r$ controls
% the number of correlated uncertainty directions modeled by $LL^\top$. Based on
% the empirical analysis, we set $r=2$ in our experiments.

This formulation compactly captures both correlated and independent pose
uncertainty. The low-rank component models coupled ambiguities across pose
dimensions, while the diagonal component accounts for residual per-parameter
uncertainty. The predicted distribution is used during training:
the covariance is supervised by a distribution loss and further converted into
joint-wise precision weights to guide the deterministic pose estimator, as
described below.

\subsection{Model Training}
\label{sec:model_training}

Following \cite{SAT-HMR}, we use Hungarian matching to associate
predicted meshes with ground-truth instances. The deterministic branch is trained
with the standard detection and reconstruction losses, including pose, shape, 3D
joint, projected 2D joint, bounding box, and detection objectives. We further add
a distribution loss and a heteroscedastic joint loss, yielding:
\begin{equation}
\begin{aligned}
\mathcal{L}_{\mathrm{total}}
&=
\lambda_{\mathrm{pose}}\mathcal{L}_{\mathrm{pose}}
+
\lambda_{\mathrm{shape}}\mathcal{L}_{\mathrm{shape}}
+
\lambda_{\mathrm{j3d}}\mathcal{L}_{\mathrm{j3d}}
\\&+
\lambda_{\mathrm{j2d}}\mathcal{L}_{\mathrm{j2d}}
\quad+
\lambda_{\mathrm{box}}\mathcal{L}_{\mathrm{box}}
+
\lambda_{\mathrm{det}}\mathcal{L}_{\mathrm{det}}
\\&+
\lambda_{\mathrm{dist}}\mathcal{L}_{\mathrm{dist}}
+
\lambda_{\mathrm{hetero}}\mathcal{L}_{\mathrm{hetero}} .
\end{aligned}
\label{eq:total_loss}
\end{equation}
Here, $\mathcal{L}_{\mathrm{box}}$ combines an $L_1$ regression term with the
generalized IoU loss~\cite{GIoU}, and $\mathcal{L}_{\mathrm{det}}$ uses focal
loss~\cite{focal_loss}. The remaining standard reconstruction losses ($\mathcal{L}_{\mathrm{pose}}, \mathcal{L}_{\mathrm{shape}}, \mathcal{L}_{\mathrm{j3d}}, \mathcal{L}_{\mathrm{j2d}}$) are
implemented with $L_1$ loss.

\paragraph{Distribution Loss.}
We train the covariance branch with a Gaussian negative log-likelihood loss.
Given the matched ground-truth 6D pose $\mu^{\mathrm{gt}}$, we define
\begin{equation}
\mathcal{L}_{\mathrm{dist}}
=
-\log
\mathcal{N}
\left(
\mu^{\mathrm{gt}};
\mathrm{sg}[\mu],\Sigma
\right),
\label{eq:nll}
\end{equation}
where $\mathrm{sg}[\cdot]$ denotes stop-gradient. This loss models
the residual $\mu^{\mathrm{gt}}-\mathrm{sg}[\mu]$ as a zero-mean Gaussian with
covariance $\Sigma$. Detaching $\mu$ prevents the likelihood term from directly
updating the deterministic pose branch, so $\mathcal{L}_{\mathrm{dist}}$ trains
the covariance to explain residual pose uncertainty.

\paragraph{Heteroscedastic Joint Loss.}
To make the learned uncertainty guide the deterministic estimator, we convert the
predicted covariance into joint-wise precision-inspired weights. For each joint
$j$, we estimate a joint-level variance by averaging the corresponding diagonal
entries of $\Sigma$, convert it to an inverse-variance score, and normalize the
weights within each person:
\begin{equation}
\sigma_j^2
=
\frac{1}{|\mathcal{I}_j|}
\sum_{i\in\mathcal{I}_j}
\mathrm{diag}(\Sigma)_i,
\qquad
w_j
=
\frac{1/\sigma_j^2}
{\frac{1}{J}\sum_{k=1}^{J}1/\sigma_k^2}.
\label{eq:joint_weight}
\end{equation}
Here, $\mathcal{I}_j$ denotes the 6D pose dimensions associated with joint
$j$. The variance $\sigma_j^2$ represents the covariance-derived residual
uncertainty of joint $j$, and $w_j$ is a normalized relative reliability weight.
This weighting is not intended as hard-example mining. In classification losses
such as focal loss~\cite{focal_loss}, a hard example still has a well-defined
class label, so increasing its weight reinforces an informative supervision
signal. In monocular HMR, however, high uncertainty often indicates that the
current image does not sufficiently constrain the 3D configuration of a joint,
due to occlusion, blur, truncation, or projection ambiguity. Directly emphasizing
such joints may force a deterministic regressor to fit ambiguous residuals and
introduce noisy gradients. We therefore use the predicted uncertainty as a
reliability signal: low-uncertainty joints serve as more trustworthy anchors for
regression, while high-uncertainty joints are down-weighted in the supervision. In this way, the model can focus the heteroscedastic joint loss
on better-constrained observations while still learning from ambiguous joints
through the remaining training losses.

The detached weights are applied to the root-relative 3D joint loss. Let
$\bar{\mathbf{J}}^{\mathrm{pred}}_j =
\mathbf{J}^{\mathrm{pred}}_j -
\mathbf{J}^{\mathrm{pred}}_{\mathrm{root}}$
and
$\bar{\mathbf{J}}^{\mathrm{gt}}_j =
\mathbf{J}^{\mathrm{gt}}_j -
\mathbf{J}^{\mathrm{gt}}_{\mathrm{root}}$, where $\mathbf{J}_{\mathrm{root}}$ denotes the pelvis joint. We define:
\begin{equation}
\mathcal{L}_{\mathrm{hetero}}
=
\frac{1}{J}
\sum_{j=1}^{J}
\mathrm{sg}[w_j]
\left\|
\bar{\mathbf{J}}^{\mathrm{pred}}_j
-
\bar{\mathbf{J}}^{\mathrm{gt}}_j
\right\|_1 .
\label{eq:hetero}
\end{equation}
Since $w_j$ is detached, the covariance branch cannot reduce this loss by simply
inflating uncertainty. Note that we do not replace the standard 3D joint loss
$\mathcal{L}_{\mathrm{j3d}}$ with $\mathcal{L}_{\mathrm{hetero}}$; instead,
$\mathcal{L}_{\mathrm{hetero}}$ is used as an auxiliary loss. This ensures that
all joints remain supervised by the uniform 3D joint loss, while the
heteroscedastic term emphasizes more reliable joints according to the predicted
uncertainty.

% The covariance is learned through the Gaussian residual
% likelihood in Eq.~\ref{eq:nll}, while Eq.~\ref{eq:hetero} uses it only as a
% one-way relative weighting signal. The normalization keeps the average joint
% weight unchanged within each person, so the loss redistributes gradients across
% joints rather than changing the overall scale.
\section{Experiments}

\begin{table*}[!t]
\centering
\caption{Quantitative comparison on two sports benchmarks (VIP-HockeyPose-v2, WorldPose) and the general-scene 3DPW. \textbf{Best} and \underline{second best} are marked per column. ``Frames'' is the number of input frames; for PSVT we list both its single-frame and its best 9-frame setting. Time denotes per-frame inference time in milliseconds. $^\dag$: 3DPW fine-tuning protocol not specified in~\cite{DepthHMR}.}
\label{tab:quant_result}
\setlength{\tabcolsep}{5pt}
\definecolor{oursrow}{gray}{0.92}
\resizebox{\textwidth}{!}{%
\begin{tabular}{l cccc ccc !{\color{black!12}\vrule} ccc !{\color{black!12}\vrule} ccc}
\toprule
\multirow{2}{*}{Method} & \multirow{2}{*}{Res.} & \multirow{2}{*}{Time (ms)} & \multirow{2}{*}{Backbone} & \multirow{2}{*}{Frames}
& \multicolumn{3}{c}{VIP-HockeyPose-v2~\cite{SSE}}
& \multicolumn{3}{c}{WorldPose~\cite{WorldPose}}
& \multicolumn{3}{c}{3DPW~\cite{3DPW}} \\
\cmidrule(lr){6-8} \cmidrule(lr){9-11} \cmidrule(lr){12-14}
& & & & & MPJPE\,$\downarrow$ & PA-MPJPE\,$\downarrow$ & PVE\,$\downarrow$
   & MPJPE\,$\downarrow$ & PA-MPJPE\,$\downarrow$ & PVE\,$\downarrow$
   & MPJPE\,$\downarrow$ & PA-MPJPE\,$\downarrow$ & PVE\,$\downarrow$ \\
\midrule
ROMP~\cite{ROMP}          & 512 &25.7& HRNet-32 & 1 & 58.6 & 35.5 & 77.4 & 74.0 & 40.7 & 104.8 & 76.7 & 47.3 & 93.4 \\
BEV~\cite{BEV}            & 512 &38.1& HRNet-32 & 1 & 52.7 & 31.0 & 70.4 & 67.9 & 38.6 & 94.0  & 78.5 & 46.9 & 92.3 \\
PSVT (single frame)~\cite{PSVT}          & 512 &--& HRNet-32 & 1 & --   & --   & --   & --   & --   & --    & 75.5 & 45.7 & 84.9 \\
PSVT~\cite{PSVT}          & 512 &--& HRNet-32 & 9 & --   & --   & --   & --   & --   & --    & 73.1 & 43.5 & 84.0 \\
Multi-HMR~\cite{MultiHMR} & 896 &64.8& ViT-L    & 1 & 49.6 & 30.0 & 68.0 & 59.0 & 36.3 & 85.7  & \textbf{61.4} & 41.7 & 75.9 \\
DepthHMR$^\dag$~\cite{DepthHMR}  & 644 &--& ViT-L    & 1 & --   & --   & --   & --   & --   & --    & --   & 41.6 & \textbf{71.3} \\
SAT-HMR~\cite{SAT-HMR}    & 644 &28.1& ViT-B    & 1 & 44.5 & 26.5 & 62.9 & 54.7 & 30.2 & 82.6 & 63.6 & 41.6 & 73.7 \\
\midrule
\rowcolor{oursrow}
UT-HMR (single-frame) & 644 &28.1& ViT-B & 1 & \underline{43.9} & \underline{26.2} & \underline{61.7} & \underline{54.0} & \underline{29.6} & \underline{81.5} & 62.1 & \underline{40.9} & 72.2\\
\rowcolor{oursrow}
\textbf{UT-HMR (Ours)} & 644 &39.5& ViT-B & 3 & \textbf{42.4} & \textbf{25.5} & \textbf{59.4} & \textbf{53.1} & \textbf{29.1} & \textbf{80.7} & \underline{61.7} & \textbf{40.2} & \underline{71.7} \\
\bottomrule
\end{tabular}
}
\end{table*}

\subsection{Experimental Setup}
\label{sec:exp_setup}

\subsubsection{Datasets and Metrics}
\label{sec:datasets_metrics}

We initialize UT-HMR from the SAT-HMR \cite{SAT-HMR} checkpoint, which is
pre-trained on AGORA~\cite{AGORA}, BEDLAM~\cite{BEDLAM}, COCO~\cite{COCO},
CrowdPose~\cite{CrowdPose}, MPII~\cite{MPII}, and Human3.6M~\cite{H3.6M}. We
fine-tune and evaluate the model on three datasets: VIP-HockeyPose-v2~\cite{SSE},
WorldPose~\cite{WorldPose}, and 3DPW~\cite{3DPW}.

VIP-HockeyPose-v2 is an ice hockey HMR dataset with frames sampled from 24 NHL
games. Unlike the original version~\cite{SSE}, v2 uses the more recent
SAM-3D-Body~\cite{SAM-3D-Body} to obtain higher-quality pseudo annotations.
WorldPose is a soccer HMR dataset collected from the 2022 FIFA World Cup, with
pseudo ground truth obtained by triangulating multi-view 2D keypoints from
calibrated cameras and optimizing SMPL parameters. We additionally evaluate on
3DPW, a standard in-the-wild HMR benchmark, to assess generalization beyond
sports.

We report MPJPE, PA-MPJPE, and PVE on all datasets. MPJPE measures 3D joint
reconstruction error in the original coordinate space, while PA-MPJPE measures
the error after Procrustes alignment, removing global translation, rotation,
and scale differences. PVE measures the average vertex error of the recovered
mesh. We also report average per-frame inference time in milliseconds,
measured on a single NVIDIA RTX 6000 Ada GPU.

% VIP-HockeyPose-v2 is an ice hockey HMR dataset with frames sampled from 24 NHL
% games. Different from the original version \cite{SSE}, v2 uses more recent
% SAM-3D-Body~\cite{SAM-3D-Body} to obtain higher-quality pseudo annotations. WorldPose is a soccer HMR dataset collected from
% the 2022 FIFA World Cup, with pseudo ground truth obtained by triangulating
% multi-view 2D keypoints from calibrated cameras and fitting SMPL parameters
% through optimization. We further evaluate on 3DPW, a standard in-the-wild HMR
% benchmark, to assess generalization beyond sports scenes.

% We report MPJPE, PA-MPJPE, and PVE on all datasets. MPJPE measures the 3D joint reconstruction error in the original coordinate space. PA-MPJPE evaluates the joint error after Procrustes alignment, which removes global translation, rotation, and scale differences between the prediction and ground truth. PVE measures the average vertex error of the recovered mesh. We also report the
% average per-frame inference time in milliseconds, measured on a single NVIDIA
% RTX 6000 Ada GPU.

\subsubsection{Implementation Details}
\label{sec:impl_details}

Following SAT-HMR~\cite{SAT-HMR}, we adopt a base input resolution of 644 and
use ViT-B/14 as the backbone. UT-HMR uses clips of $T{=}3$ frames: each frame is encoded independently, the
token memories are concatenated, and the center-frame queries cross-attend to
the multi-frame memory through the DETR-style decoder.

% The probabilistic branch is attached to the final decoder-layer query feature.
% In parallel with the deterministic pose head, it predicts a low-rank-plus-diagonal
% Gaussian covariance $\Sigma=LL^\top+\operatorname{diag}(d)$, where
% $L\in\mathbb{R}^{144\times r}$, $d\in\mathbb{R}^{144}$ and $r=2$ in our experiments.
%  The covariance is
% supervised by the distribution loss in Eq.~\ref{eq:nll} and converted into
% detached per-joint precision weights for the dynamic joint loss in
% Eq.~\ref{eq:hetero}. At inference time, UT-HMR directly uses the deterministic
% pose prediction.

We initialize UT-HMR from the SAT-HMR checkpoint and fine-tune the decoder,
prediction heads, and the last three ViT blocks. The newly added low-rank and
diagonal covariance heads are implemented as MLPs and trained from scratch. We
use AdamW with a learning rate of $2{\times}10^{-5}$, weight decay $10^{-4}$,
batch size 8, and gradient clipping at 1.0. We fine-tune for 50 epochs on
VIP-HockeyPose-v2 and WorldPose to account for the sports-domain shift, and
20 epochs on 3DPW. All experiments use a single NVIDIA RTX 6000 Ada GPU.

For fair comparison, all competing methods, including the SAT-HMR baseline,
are fine-tuned on each benchmark's training split, either by us using their
official code, or as reported in their papers. The sole exception is DepthHMR \cite{DepthHMR} (code unavailable), whose 3DPW fine-tuning protocol is unspecified; we include its reported numbers as-is.

% We initialize UT-HMR from the SAT-HMR checkpoint and fine-tune the decoder,
% prediction heads, and the last three blocks of the ViT backbone. The newly
% added low-rank and diagonal covariance heads are implemented as MLPs and trained
% from scratch. We use AdamW with learning
% rate $2{\times}10^{-5}$, weight decay $10^{-4}$, batch size 8, gradient clipping
% at 1.0. We fine-tune for 50 epochs on
% VIP-HockeyPose-v2 and WorldPose due to their sports-domain shift, and for 20 epochs
% on 3DPW. All experiments are conducted on a single NVIDIA RTX 6000 Ada GPU.

% \textcolor{blue}{For a fair comparison, all competing methods are also fine-tuned on the training split of each target
% benchmark from their released pre-trained weights using their official training code and default
% schedules, rather than being evaluated off-the-shelf.}

\subsection{Overall Performance}

\subsubsection{Quantitative Results}
\label{sec:quant_results}
Tab.~\ref{tab:quant_result} compares UT-HMR with recent one-stage multi-person
HMR methods. Across all three benchmarks, UT-HMR consistently improves over
the SAT-HMR baseline with only moderate computational overhead, while
remaining much faster than larger ViT-L-based methods. On both
VIP-HockeyPose-v2 and WorldPose, UT-HMR achieves the best performance on all
three metrics among the tested models; we do not compare with the results
reported in~\cite{SSE}, since VIP-HockeyPose-v2 uses updated annotations that
differ from the original version. On the general-scene 3DPW benchmark, UT-HMR remains competitive with the
strongest prior methods despite using a lighter backbone and a lower input
resolution than the top-performing competitors. Overall, UT-HMR
improves robustness in challenging sports scenarios while generalizing well
to in-the-wild scenes.

Since most existing one-stage HMR methods operate on single
frames, we additionally train a single-frame ($T{=}1$) variant of UT-HMR for
a fairer comparison. As shown by Tab. \ref{tab:quant_result}, this variant still
outperforms all competing single-frame methods on both sports benchmarks and
remains competitive on 3DPW against methods with substantially larger
backbones, indicating that the gains do not arise solely from using multiple
frames. Among multi-frame methods, PSVT reports results only on 3DPW, where
our three-frame UT-HMR also outperforms its best nine-frame setting.

% \begin{table}[t]\centering\small
% \begin{tabular*}{\columnwidth}{@{\extracolsep{\fill}}lccc@{}}
% \toprule
% Accel (mm/s$^2$)\,$\downarrow$ & SAT-HMR & UT-HMR & $\Delta$ \\
% \midrule
% VIP-HockeyPose-v2 & 51.8 & \textbf{44.7} & $-13.7\%$ \\
% WorldPose         & 66.1 & \textbf{33.4} & $-49.5\%$ \\
% 3DPW              & 20.8 & \textbf{9.8}  & $-53.1\%$ \\
% \bottomrule
% \end{tabular*}
% \caption{Temporal smoothness measured by acceleration error. Temporal early fusion roughly halves the
% jitter on WorldPose and 3DPW relative to the single-frame baseline.}
% \label{tab:accel}
% \end{table}

\subsubsection{Qualitative Results}
\label{sec:qual_results}

\begin{figure*}[!htbp]
\centering
\begin{subfigure}{0.32\textwidth}
\centering
\includegraphics[width=\linewidth]{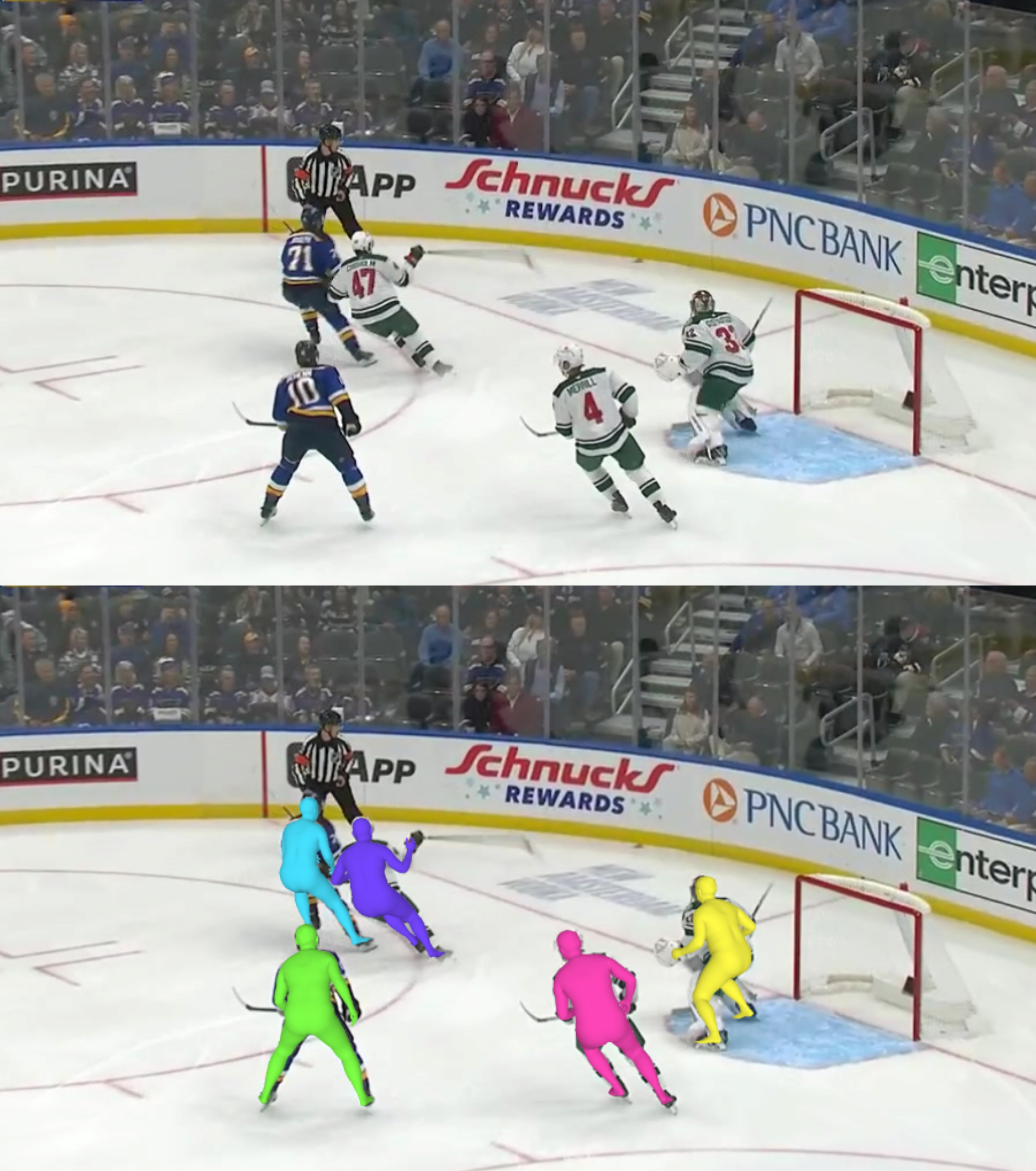}
\caption{VIP-HockeyPose-v2}
\label{fig:qual_vip}
\end{subfigure}
\hfill
\begin{subfigure}{0.32\textwidth}
\centering
\includegraphics[width=\linewidth]{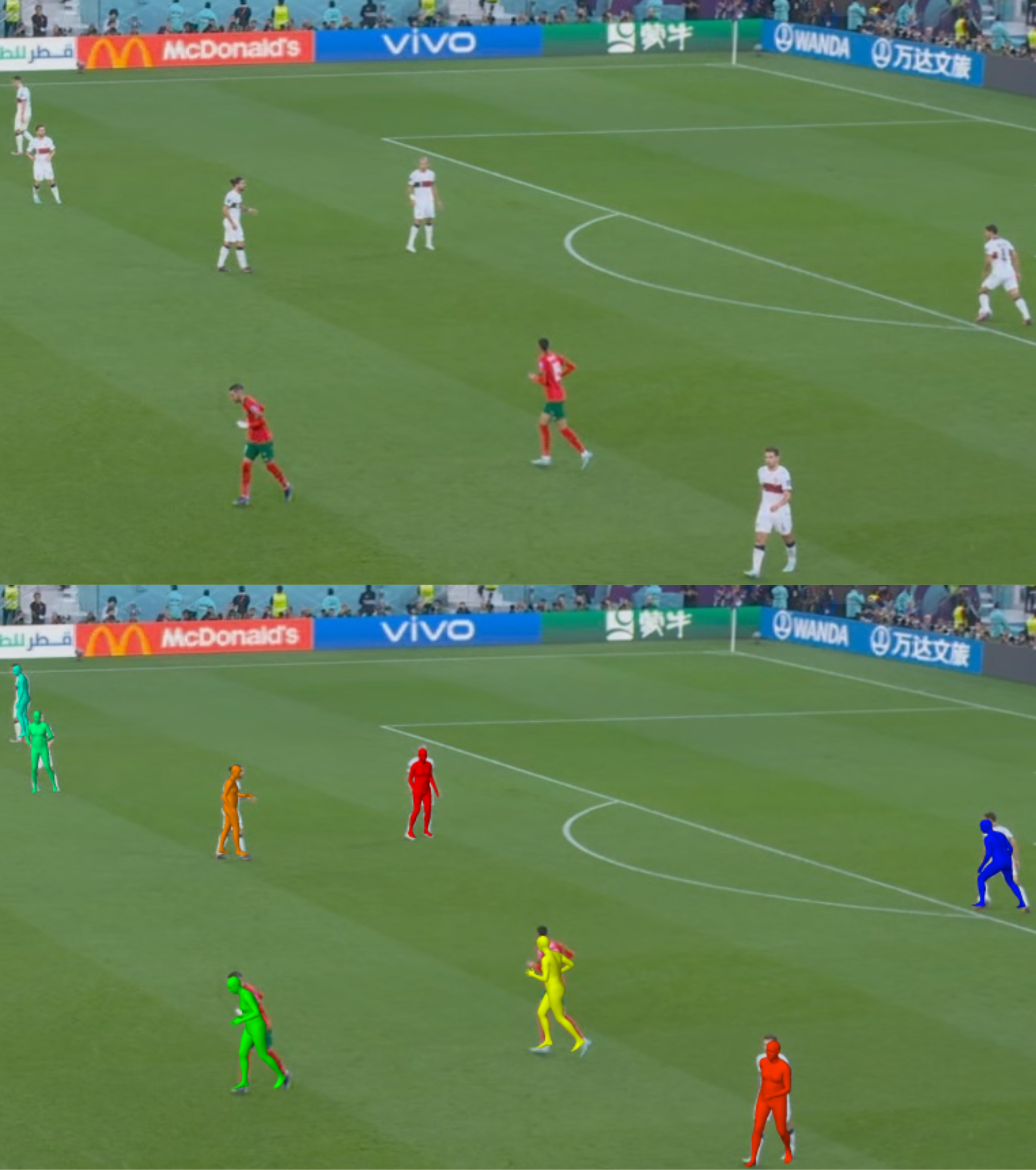}
\caption{WorldPose}
\label{fig:qual_worldpose}
\end{subfigure}
\hfill
\begin{subfigure}{0.32\textwidth}
\centering
\includegraphics[width=\linewidth]{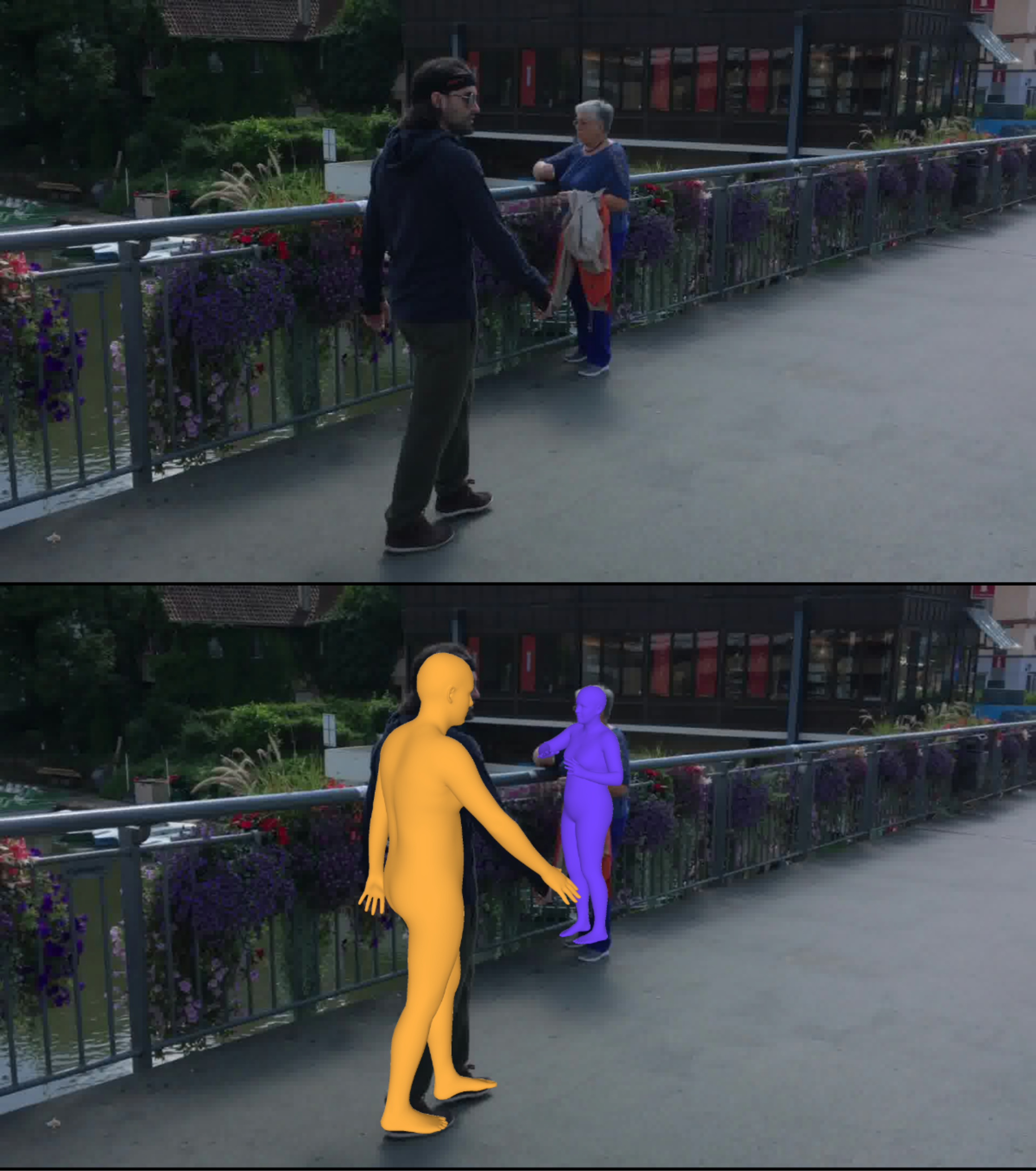}
\caption{3DPW}
\label{fig:qual_3dpw}
\end{subfigure}

\caption{
\textbf{Qualitative results on different benchmarks.}
UT-HMR reconstructions on (a) VIP-HockeyPose-v2~\cite{SSE},
(b) WorldPose~\cite{WorldPose}, and (c) 3DPW~\cite{3DPW} are overlaid on input
frames. Note that referees
are excluded from evaluation on VIP-HockeyPose-v2 because ground-truth meshes are
not available for them.}
\label{fig:qualitative}
\end{figure*}

We visualize qualitative results of UT-HMR across different datasets in Fig.~\ref{fig:qualitative}. The recovered meshes are overlaid on the corresponding input frames.

As shown in Fig.~\ref{fig:qualitative}, UT-HMR produces stable and plausible
mesh reconstructions across both sports and general-scene datasets. On
challenging ice hockey and soccer frames, it remains robust to small person
scale, occlusion, and motion blur, while also generalizing well to in-the-wild
scenes on 3DPW.

\begin{figure}[htbp]
\centering
\begin{subfigure}{0.15\textwidth}
\centering
\includegraphics[width=\linewidth]{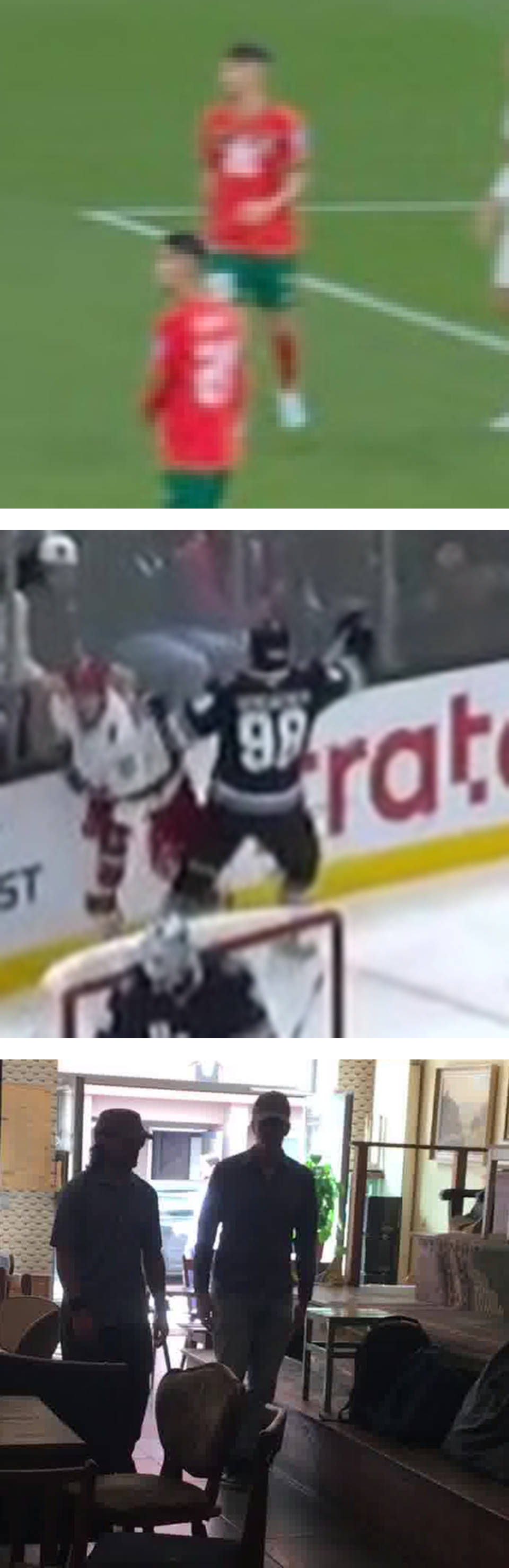}
\caption{Frame}
\label{fig:comparison_frame}
\end{subfigure}
\hfill
\begin{subfigure}{0.15\textwidth}
\centering
\includegraphics[width=\linewidth]{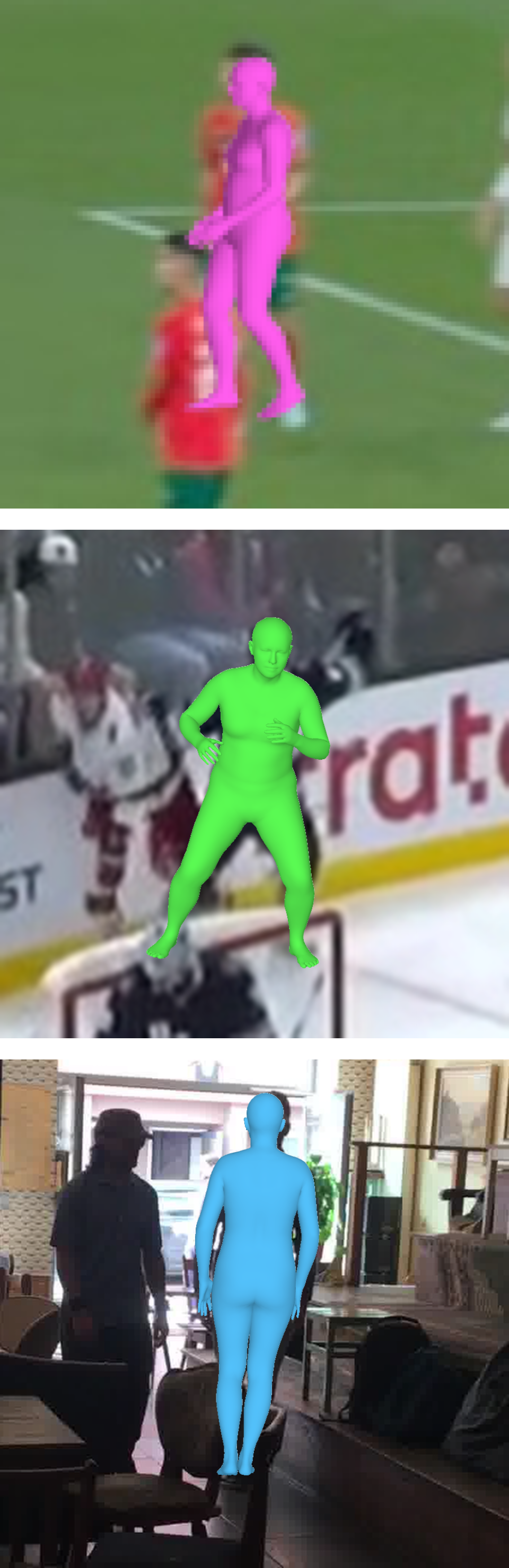}
\caption{SAT-HMR}
\label{fig:comparison_baseline}
\end{subfigure}
\hfill
\begin{subfigure}{0.15\textwidth}
\centering
\includegraphics[width=\linewidth]{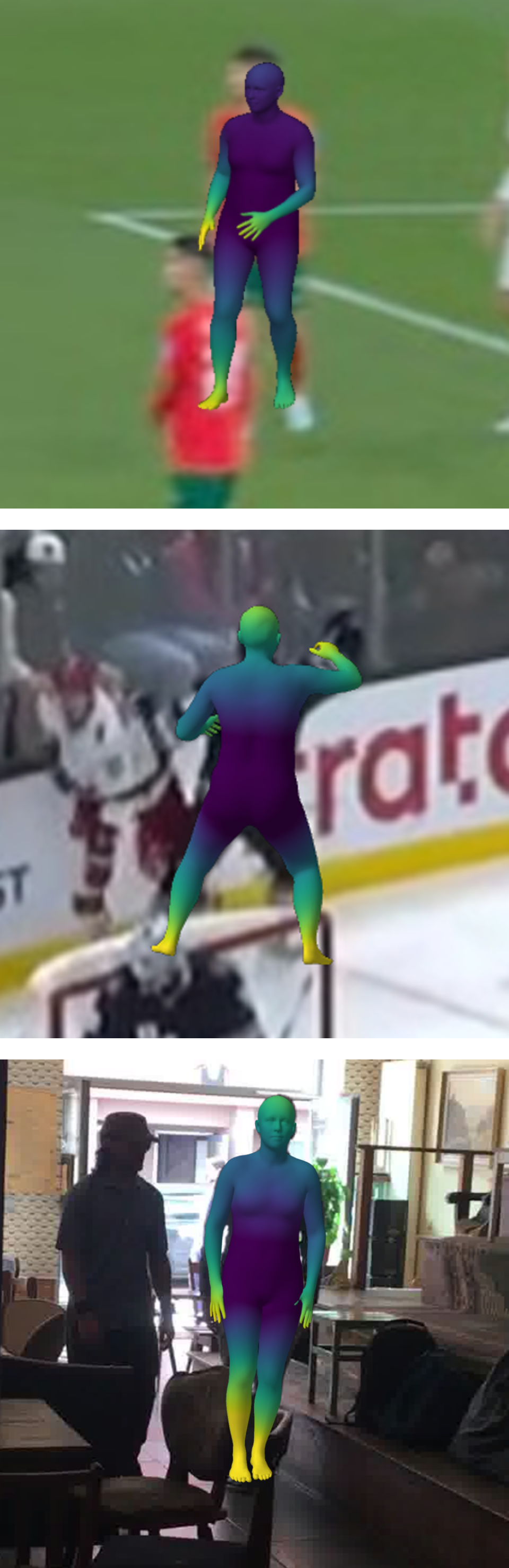}
\caption{UT-HMR}
\label{fig:comparison_ours}
\end{subfigure}

\vspace{1mm}
\centering
{\scriptsize Low uncertainty}
\hspace{1mm}
\includegraphics[width=0.25\textwidth,height=1.2mm]{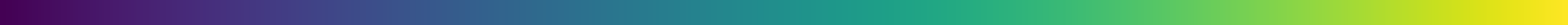}
\hspace{1mm}
{\scriptsize High uncertainty}

\caption{
\textbf{Qualitative comparison with SAT-HMR.}
Meshes predicted by SAT-HMR~\cite{SAT-HMR} and UT-HMR are overlaid on
representative challenging crops. Lighter colors indicate higher predicted
uncertainty. Please zoom in for details.
}
\label{fig:comparison}
\end{figure}

To further illustrate the improvement, Fig.~\ref{fig:comparison} compares
UT-HMR with SAT-HMR~\cite{SAT-HMR} on representative challenging frames.
UT-HMR reduces failure cases in sports scenarios, particularly for unclear or
occluded players, while maintaining strong generalization on 3DPW. The
visualized uncertainty is higher around limbs, especially under occlusion or
weak visual evidence, suggesting that the learned uncertainty captures
meaningful pose ambiguity.

% As shown in Fig.~\ref{fig:qualitative}, UT-HMR produces stable and plausible
% mesh reconstructions across both sports and general-scene datasets. In
% challenging ice hockey and soccer frames, the model remains robust to small
% person scale, occlusion, and motion blur, while also generalizing well to
% in-the-wild scenes on 3DPW. 

% To further highlight the improvement over the baseline, we compare UT-HMR with
% SAT-HMR~\cite{SAT-HMR} on representative challenging frames in
% Fig.~\ref{fig:comparison}. Compared with
% SAT-HMR, UT-HMR reduces failure cases in sports scenarios, especially for
% unclear or occluded players. It also generalizes well to the general-purpose 3DPW
% dataset, indicating that the proposed design is not limited to sports-specific
% scenarios. The predicted uncertainty is further visualized on the mesh, where
% higher uncertainty tends to appear around limbs, especially those affected by
% occlusion or weak image evidence. This suggests that the learned uncertainty
% captures meaningful pose ambiguity.

\subsection{Ablation Studies}

\subsubsection{Effectiveness of Proposed Modifications}
\label{abla_all}

We evaluate the proposed components on VIP-HockeyPose-v2, with results summarized
in Tab.~\ref{tab:abla_all}. Both temporal early fusion and pose uncertainty
modeling improve over the baseline, while their combination achieves the best
performance, demonstrating their effectiveness.

\begin{figure}
    \centering
    \includegraphics[width=\linewidth]{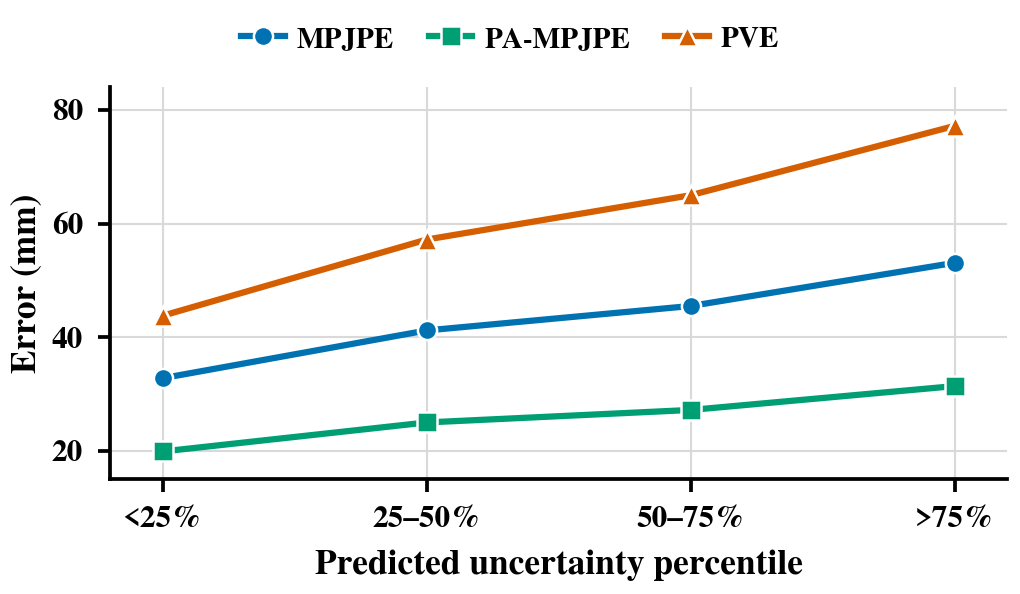}
    \caption{
Reconstruction error versus predicted joint-uncertainty percentile on
VIP-HockeyPose-v2.
}
\label{fig:uncertainty_bin}
\end{figure}

We further assess whether the learned distribution captures meaningful pose
uncertainty through an uncertainty-stratified analysis. For each person, we compute joint-level uncertainty as the marginal variance
$\sigma_j^2$ averaged over the six dimensions of joint $j$'s 6D rotation, and
obtain an instance-level score by averaging over all joints. We then divide instances
into four equal-size percentile groups. As shown in Fig.~\ref{fig:uncertainty_bin},
MPJPE, PA-MPJPE, and PVE all increase monotonically with predicted uncertainty,
indicating that the learned covariance provides a meaningful reliability
ranking and supports its use for joint reweighting.

% We evaluate the proposed components on VIP-HockeyPose-v2, with results summarized in
% Tab.~\ref{tab:abla_all}. Both temporal early fusion and pose uncertainty
% modeling improve over the baseline, demonstrating the effectiveness of the two proposed components. Moreover, combining the two components achieves the
% best performance, showing that temporal context and uncertainty modeling are
% complementary.

% We further examine whether the learned distribution provides meaningful pose
% uncertainty through the uncertainty-stratified analysis. For each person, we compute joint-level
% uncertainty using the covariance-derived variance $\sigma_j^2$, which averages the marginal variances of the six 6D
% pose dimensions associated with joint $j$. We then obtain an instance-level
% uncertainty score by averaging $\sigma_j^2$ over all joints, and divide all
% instances into four equal-size percentile groups. As shown in
% Fig.~\ref{fig:uncertainty_bin}, MPJPE, PA-MPJPE, and PVE all increase
% monotonically with predicted uncertainty, suggesting that the learned covariance
% provides a meaningful reliability ranking. This supports its use as a training
% signal for joint reweighting.

\begin{table}[t]
    \centering
    \setlength{\tabcolsep}{3pt}
    \begin{tabular}{ccccc}
    \hline
    Temporal & Uncertainty & MPJPE $\downarrow$ & PA-MPJPE $\downarrow$ & PVE $\downarrow$ \\
    \hline
    -- & -- & 44.5 & 26.5 & 62.9 \\
    \checkmark & -- & 43.5 & 26.1 & 61.0 \\
    -- & \checkmark & 43.9 & 26.2 & 61.7 \\
    \checkmark & \checkmark & \textbf{42.4} & \textbf{25.5} & \textbf{59.4} \\
    \hline
    \end{tabular}
    \caption{Ablation of temporal early fusion and pose uncertainty modeling on VIP-HockeyPose-v2. The SAT-HMR baseline (first row) is fine-tuned on the same data as UT-HMR.}
    \label{tab:abla_all}
\end{table}

% \begin{table}[t]
% \centering
% \setlength{\tabcolsep}{5pt}
% \begin{tabular}{lccc}
% \hline
% Uncertainty percentile & MPJPE$\downarrow$ & PA-MPJPE$\downarrow$ & PVE$\downarrow$ \\
% \hline
% $<25\%$     & 32.8 & 19.9 & 43.8 \\
% $25$--$50\%$  & 41.2 & 25.0 & 57.2 \\
% $50$--$75\%$  & 45.5 & 27.2 & 65.0 \\
% $>75\%$     & 53.1 & 31.4 & 77.2 \\
% \hline
% \end{tabular}
% \caption{
% Uncertainty-error stratification on VIP-HockeyPose. Instances are divided into
% four bins by average predicted joint uncertainty. Higher uncertainty consistently
% corresponds to larger reconstruction errors.
% }
% \label{tab:uncertainty_bin}
% \end{table}

% Preamble:
% \usepackage{pgfplots}
% \pgfplotsset{compat=1.18}

\subsubsection{Hyperparameter Sensitivity}
\label{sec:abla_hyper}

We analyze the sensitivity of UT-HMR to three key design choices: the temporal range used in early fusion, the covariance rank $r$ of the uncertainty branch, and the granularity of uncertainty-based reweighting. The results are summarized in Tab.~\ref{tab:hyper}. As shown in the table, a short temporal window improves over the
single-frame setting, confirming that neighboring frames help resolve ambiguity. Increasing the range beyond $\pm1$ brings no further gain and may
slightly degrade performance, likely due to redundant or less aligned distant
frames. For the covariance rank, introducing a low-rank factor improves over the
diagonal-only ($r=0$) setting, showing the benefit of modeling correlated pose
uncertainty. However, larger ranks degrade performance, suggesting that pose
uncertainty is concentrated in a few dominant directions. Therefore, a small rank
is sufficient and avoids noisy extra dimensions. We also find that
per-joint weighting outperforms per-person weighting, suggesting that ambiguity
often affects specific joints rather than the whole person. Based on these
results, we use temporal range $\pm1$, covariance rank $r=2$, and per-joint
uncertainty weighting in the final model.

% We analyze the sensitivity of UT-HMR to three key design choices: the temporal range used in early fusion, the covariance rank $r$ of the uncertainty branch, and the granularity of uncertainty-based reweighting. The results are summarized in Tab.~\ref{tab:hyper}. As shown in the table, a short temporal window improves over the
% single-frame setting, confirming that neighboring frames help resolve ambiguity. Increasing the range beyond $\pm1$ brings no further gain and may
% slightly degrade performance, likely due to redundant or less aligned distant
% frames. For the covariance rank, introducing a low-rank factor improves over the
% diagonal-only ($r=0$) setting, showing the benefit of modeling correlated pose
% uncertainty. However, larger ranks degrade performance, suggesting that pose
% uncertainty is concentrated in a few dominant directions. Therefore, a small rank
% is sufficient and avoids noisy extra dimensions. We also find that
% per-joint weighting outperforms per-person weighting, suggesting that ambiguity
% often affects specific joints rather than the whole person. Based on these
% results, we use temporal range $\pm1$, covariance rank $r=2$, and per-joint
% uncertainty weighting in the final model.

\begin{table}[t]
\centering
\setlength{\tabcolsep}{4pt}
\begin{tabular}{llccc}
\hline
Param. & Value & MPJPE$\downarrow$ & PA-MPJPE$\downarrow$ & PVE$\downarrow$\\
\hline
Range & $\pm0$ & 43.9 & 26.2 & 61.7\\
Range & $\pm1$ & \textbf{42.4} & \textbf{25.5} & \textbf{59.4}\\
Range & $\pm2$ & 42.4 & 25.5 & 59.7\\
% Range & $\pm3$ & 43.8 & 26.0 & 61.2\\
% Range & $\pm5$ & 43.7 & 26.1 & 61.1\\
\hline
Rank $r$ & 0  & 43.1 & 25.6 & 60.6\\
Rank $r$ & 2  & \textbf{42.4} & \textbf{25.5} & \textbf{59.4}\\
Rank $r$ & 4  & 42.7 & 25.6 & 60.0\\
% Rank $r$ & 8  & 42.6 & 25.6 & 59.8\\
% Rank $r$ & 16 & 43.1 & 25.6 & 60.6\\
\hline
Weighting & Per-person & 42.8 & 25.5 & 60.1\\
Weighting & Per-joint  & \textbf{42.4} & \textbf{25.5} & \textbf{59.4}\\
\hline
\end{tabular}
\caption{
Hyperparameter sensitivity on VIP-HockeyPose-v2. Note that $r{=}0$ denotes a diagonal covariance, which still models uncertainty and thus differs from the temporal-only baseline.
}
\label{tab:hyper}
\end{table}

\subsubsection{Effectiveness of Heteroscedastic Joint Loss}
\label{sec:abl_weighting}

To evaluate the effectiveness of our heteroscedastic weighting, we compare
several variants that differ only in the per-joint weight $w_j$: uniform
weighting, error-based weighting, and weights derived from the predicted
position uncertainty. The error-based variant follows a focal-style strategy that
up-weights joints with larger prediction errors, while the uncertainty-based
variants use the predicted uncertainty to either up-weight or down-weight uncertain joints.
Results are summarized in Tab.~\ref{tab:weighting}.

\begin{table}[t]
\centering
\setlength{\tabcolsep}{3pt}
\begin{tabular}{lccc}
\toprule
Weighting $w_j$ & MPJPE$\downarrow$ & PA-MPJPE$\downarrow$ & PVE$\downarrow$\\
\midrule
Uniform                         & 43.5 & 26.1 & 61.0\\
Error-based                     & 43.4 & 26.0 & 60.9\\
High-unc. weighted              & 42.8 & 25.6 & 59.9\\
Low-unc. weighted (ours)        & \textbf{42.4} & \textbf{25.5} & \textbf{59.4}\\
\bottomrule
\end{tabular}
\caption{
Effect of the joint-weighting source on VIP-HockeyPose-v2. ``High-unc.'' assigns
larger weights to uncertain joints, while ``Low-unc.'' emphasizes reliable joints.
}
\label{tab:weighting}
\end{table}

As shown in Tab.~\ref{tab:weighting}, error-based weighting performs similarly
to the uniform baseline, while uncertainty-based weighting consistently improves
the results. Up-weighting uncertain joints is better than uniform or error-based
weighting, but remains inferior to our down-weighting strategy. This supports
our hypothesis in Sec.~\ref{sec:model_training} that emphasizing reliable joints
and down-weighting ambiguous ones is more effective for training.

\section{Conclusion}

In this paper, we presented UT-HMR, a temporal and uncertainty-guided framework
for one-stage multi-person human mesh recovery. UT-HMR combines temporal early
fusion with low-rank-plus-diagonal Gaussian pose uncertainty modeling and uses the predicted
covariance for heteroscedastic joint supervision. Experiments on VIP-HockeyPose-v2, WorldPose, and 3DPW demonstrate consistent
improvements over the baseline, while remaining competitive with larger models
despite using a lighter backbone.

Beyond accuracy, the predicted covariance is a usable output in
itself: it provides a per-person reliability
signal at negligible inference cost. For tracking, it can weight data association
under weak observations; for action analysis, the joint-wise uncertainty can
down-weight unreliable joints when aggregating pose sequences. We leave a
quantitative study of such applications to future work.

% In this paper, we presented UT-HMR, a temporal and uncertainty-guided framework
% for one-stage multi-person human mesh recovery. UT-HMR combines temporal early
% fusion with low-rank Gaussian pose uncertainty modeling, and uses the predicted
% covariance for heteroscedastic joint supervision. Experiments on VIP-HockeyPose-v2,
% WorldPose, and 3DPW show that UT-HMR improves over the baseline and remains
% competitive with a lighter backbone, demonstrating the effectiveness of the
% proposed framework.

% \textcolor{blue}{Beyond accuracy, the predicted covariance is a usable output in itself: since reconstruction error grows
% monotonically with it, it can act as a reliability signal for downstream tasks such as tracking or action
% analysis, at no extra inference cost. We leave a quantitative study of such applications to future work.}
{
    \small
    \bibliographystyle{ieeenat_fullname}
    \bibliography{main}
}

\end{document}

\maketitle
\appendix
\section{Temporal Evaluation}
To assess whether the improvements of UT-HMR are consistent over time rather
than limited to per-frame accuracy, we report the acceleration error (Accel),
a standard metric for temporal smoothness in video-based HMR.
Accel measures the mean difference between the second-order derivatives
(accelerations) of the predicted and ground-truth 3D joint trajectories, in
mm/s$^2$; lower values indicate smoother and more temporally coherent
predictions. For a fair comparison, SAT-HMR and UT-HMR are evaluated under
identical Hungarian matching to the ground-truth instances, and Accel is
computed only over frames where a person is consistently matched.

Results are summarized in Tab.~\ref{tab:accel}. UT-HMR substantially reduces
the acceleration error across all three benchmarks, roughly halving the jitter
on WorldPose and 3DPW. This
improvement indicates that our method yields temporally more
stable predictions, complementing the per-frame accuracy gains reported in the
main paper.

\begin{table}[htbp]
\centering\footnotesize
\renewcommand{\arraystretch}{0.95}
\begin{tabular*}{\columnwidth}{@{\extracolsep{\fill}}lccc@{}}
\hline
\rule{0pt}{2.2ex}Accel (mm/s$^2$) $\downarrow$ & SAT-HMR & UT-HMR (ours) & $\Delta$ \\
\hline
VIP-HockeyPose-v2 & 51.8 & \textbf{44.7} & $-$13.7\% \\
WorldPose         & 66.1 & \textbf{33.4} & $-$49.5\% \\
3DPW              & 20.8 & \textbf{9.8}  & $-$52.9\% \\
\hline
\end{tabular*}
\caption{Acceleration error (mm/s$^2$) on all three benchmarks. Our model substantially reduces temporal jitter, roughly halving the error
on WorldPose and 3DPW.}
\label{tab:accel}
\end{table}

\section{Interaction between Temporal Fusion and Uncertainty Modeling}
The main paper shows that temporal fusion and uncertainty modeling are
complementary in terms of accuracy. Here we further examine their interaction:
whether temporal evidence also improves the quality of the predicted
uncertainty.

We evaluate covariance calibration using the normalized squared Mahalanobis
distance,
\begin{equation}
\chi^2/D=(\mu^{\text{gt}}-\mu)^{\top}\Sigma^{-1}(\mu^{\text{gt}}-\mu)/D,
\qquad D=144,
\end{equation}
where $\mu$ and $\Sigma$ are the predicted 6D pose mean
and covariance of a matched person (Sec.~3.4 of the main paper), and
$\mu^{\text{gt}}$ is the corresponding ground-truth 6D pose. The metric is
averaged over all matched persons. Under a well-calibrated Gaussian, the
residual $\mu^{\text{gt}}-\mu$ is distributed according to the predicted
covariance $\Sigma$, and the expected value of $\chi^2/D$ is 1; values above 1
indicate that the model underestimates its uncertainty. We compare the full model against the variant that uses
only the center frame $[I_t]$, keeping all other components unchanged.

As reported in Tab.~\ref{tab:interaction}, incorporating neighboring frames
improves both pose accuracy and calibration, moving $\chi^2/D$ closer to 1.
This indicates that temporal evidence not only reduces pose error but also
leads to more reliable uncertainty estimates, supporting the view that the two
proposed components operate as one coherent system: temporal fusion supplies
additional evidence, and the uncertainty branch more faithfully quantifies the
ambiguity.

\begin{table}[htbp]
\centering\footnotesize
\renewcommand{\arraystretch}{0.95}
\begin{tabular*}{\columnwidth}{@{\extracolsep{\fill}}lcc@{}}
\hline
\rule{0pt}{2.2ex}Input & MPJPE $\downarrow$ & $\chi^2/D \to 1$ \\
\hline
$[I_t]$ (no temporal evidence) & 43.9 & 1.208 \\
$[I_{t-1},I_t,I_{t+1}]$        & \textbf{42.4} & \textbf{1.157} \\
\hline
\end{tabular*}
\caption{Uncertainty calibration with and without temporal evidence on
VIP-HockeyPose-v2. $\chi^2/D$ closer to 1 indicates better-calibrated
covariance.}
\label{tab:interaction}
\end{table}

\section{Additional Mesh-Level Evaluation}
\label{sec:supp_mesh}

The main paper reports PVE, the mean per-vertex distance to the ground-truth mesh, which summarizes the
whole surface in a single number. To characterize mesh quality more precisely, we additionally report its
Procrustes-aligned variant PA-PVE together with two further mesh-level measurements (shown in Tab. \ref{tab:supp_mesh}):

\begin{itemize}
    \item \textbf{Part-wise PVE.} Every vertex is assigned to the joint with the largest SMPL
    linear-blend-skinning weight, and the joints are grouped into torso, head, arms and legs. Reporting PVE
    per group shows which body parts the reconstruction error concentrates in.
    \item \textbf{Vertex accuracy.} The fraction of vertices within 10\,mm and 20\,mm of the ground truth,
    which characterizes the error distribution rather than its mean.
\end{itemize}

\begin{table*}[t]
\centering\small
\begin{tabular}{ll c c cccc cc}
\toprule
 & & & & \multicolumn{4}{c}{Part-wise PVE (mm)\,$\downarrow$} & \multicolumn{2}{c}{Vertex acc. (\%)\,$\uparrow$} \\
\cmidrule(lr){5-8}\cmidrule(lr){9-10}
Benchmark & Method & PVE\,$\downarrow$ & PA-PVE\,$\downarrow$ & torso & head & arms & legs & @10\,mm & @20\,mm \\
\midrule
\multirow{2}{*}{VIP-HockeyPose-v2} & SAT-HMR & 62.9 & 45.1 & 30.4 & 55.1 & 88.2 & 62.0 & 5.8 & 19.5 \\
 & UT-HMR & \textbf{59.4} & \textbf{42.8} & \textbf{28.8} & \textbf{52.4} & \textbf{82.7} & \textbf{59.2} & \textbf{6.1} & \textbf{20.8} \\
\midrule
\multirow{2}{*}{WorldPose} & SAT-HMR & 82.6 & 55.4 & 42.9 & 63.3 & 119.6 & 79.4 & 4.5 & 15.7 \\
 & UT-HMR & \textbf{80.7} & \textbf{54.2} & \textbf{42.0} & \textbf{62.2} & \textbf{117.2} & \textbf{76.5} & \textbf{4.8} & \textbf{16.2} \\
\midrule
\multirow{2}{*}{3DPW} & SAT-HMR & 73.7 & 53.2 & 39.2 & 65.5 & 97.0 & 78.3 & 1.6 & 9.8 \\
 & UT-HMR & \textbf{71.7} & \textbf{52.2} & \textbf{38.6} & \textbf{64.4} & \textbf{93.9} & \textbf{77.1} & \textbf{1.8} & \textbf{10.2} \\
\bottomrule
\end{tabular}
\caption{Mesh-level evaluation. In addition to the mean vertex distance (PVE) reported in the main paper,
we report its Procrustes-aligned variant, a breakdown of the error by body part, and the fraction of
vertices recovered within 10 and 20\,mm of the ground truth.}
\label{tab:supp_mesh}
\end{table*}

\subsection{Quantitative Mesh Evaluation}
As shown by Tab.~\ref{tab:supp_mesh}, two consistent patterns emerge across all three datasets. First, reconstruction
error is not uniform across the body: the arms exhibit the largest errors,
while the torso has the smallest, with roughly a three-fold difference. Second,
UT-HMR improves most in the regions where errors are concentrated, with the
largest reductions occurring on the arms, followed by the legs, while the torso changes only
modestly. This suggests that our method primarily benefits challenging body parts rather than uniformly improving the entire mesh. The accuracy-at-threshold metrics show the
same trend: the fractions of vertices within 10~mm and 20~mm of the ground-truth
surface increase consistently across all three datasets, indicating that the
improvement is not driven solely by a small number of corrected outliers.

\subsection{Qualitative Mesh Evaluation}
\label{sec:supp_mesh_qual}

Fig.~\ref{fig:supp_mesh_qual} qualitatively compares the meshes reconstructed by
SAT-HMR and UT-HMR against the ground truth on VIP-HockeyPose-v2, WorldPose and 3DPW. The examples
cover challenging cases in all three datasets. In these cases,
SAT-HMR occasionally produces incorrect limb configurations or distorted body
poses, whereas UT-HMR generally recovers more plausible poses that better match
the ground-truth meshes. These examples qualitatively demonstrate the benefit
of incorporating temporal context and uncertainty modeling for ambiguous visual
observations.

All results are generated by the one-stage models from the full input image:
all persons are detected and reconstructed in a single forward pass without
external bounding-box inputs. The per-person crops shown in
Fig.~\ref{fig:supp_mesh_qual} are used only for visualization and do not serve
as inputs.
\begin{figure*}[t]
    \centering
    \includegraphics[width=\textwidth]{figs/supp_fig_compare.png}
\caption{Qualitative comparison of reconstructed meshes. Each example shows the
input crop, the SAT-HMR baseline, UT-HMR, and the ground truth. Across all
three benchmarks, the baseline produces incorrect limb configurations, while
UT-HMR recovers poses that closely follow the ground truth.}
    \label{fig:supp_mesh_qual}
\end{figure*}

\section{Failure Case Analysis}
\label{sec:supp_failure}

To better understand when UT-HMR struggles, we analyze its reconstruction error under different
levels of observation difficulty on ice hockey. We consider three factors that can make reconstruction challenging: person scale, inter-person occlusion, and inter-frame motion.
For each matched person, these factors are computed from the ground-truth annotations, and
samples are stratified accordingly. We then compare the MPJPE of UT-HMR and SAT-HMR within
each stratum. The results are summarized in Tab.~\ref{tab:supp_failure}. Note that the analysis covers only center frames with both neighboring frames
available and persons matched to the ground truth, so the averages differ
slightly from those in Table 1 of the main paper.

\begin{itemize}
    \item \textbf{Person scale.} We measure the 2D extent of the ground-truth skeleton in pixels;
    smaller persons correspond to weaker and lower-resolution observations.

    \item \textbf{Occlusion.} We use the maximum 2D bounding-box IoU between the target person
    and any other annotated person as a proxy for inter-person occlusion.

    \item \textbf{Inter-frame motion.} We measure the mean per-joint displacement of the ground-truth between frames $t-1$ and $t+1$, characterizing the motion within temporal window.
\end{itemize}

\begin{table}[t]
\centering\small
\begin{tabular*}{\columnwidth}{@{\extracolsep{\fill}}lcc@{}}
\toprule
Stratum & UT-HMR & SAT-HMR \\
\midrule
\multicolumn{3}{l}{Person scale (quartiles, large $\rightarrow$ small)} \\
Q1 (largest)  & \textbf{36.1} & 37.6 \\
Q2            & \textbf{36.3} & 38.6 \\
Q3            & \textbf{43.9} & 45.0 \\
Q4 (smallest) & \textbf{51.0} & 52.1 \\
\midrule
\multicolumn{3}{l}{Occlusion (quartiles, low $\rightarrow$ high) } \\
Q1 (lowest)  & \textbf{37.5} & 38.5 \\
Q2  & \textbf{45.0} & 46.2 \\
Q3  & \textbf{53.1} & 54.1 \\
Q4 (highest)     & \textbf{61.9} & 62.4 \\
\midrule
\multicolumn{3}{l}{Inter-frame motion (quartiles, slow $\rightarrow$ fast)} \\
Q1 (slowest) & \textbf{35.5} & 36.9 \\
Q2           & \textbf{39.5} & 40.9 \\
Q3           & \textbf{42.9} & 44.2 \\
Q4 (fastest) & \textbf{48.9} & 50.0 \\
\bottomrule
\end{tabular*}
\caption{MPJPE (mm) under different levels of observation difficulty on VIP-HockeyPose-v2.
Samples are stratified by ground-truth person scale, inter-person occlusion, and inter-frame
motion.}
\label{tab:supp_failure}
\end{table}

\subsection{Performance under Challenging Conditions}
Tab.~\ref{tab:supp_failure} shows that all three factors contribute to larger
reconstruction errors, with MPJPE increasing consistently as the observation
becomes more challenging. Among them, occlusion has the strongest impact:
UT-HMR reaches the highest MPJPE of $61.9$~mm under the highest occlusion level,
compared with $37.5$~mm for persons with the lowest occlusion. In contrast, although person scale
and inter-frame motion also degrade performance, their worst-case errors
remain lower than that under highest occlusion.

A similar trend can be observed in the improvement over SAT-HMR. Although
UT-HMR consistently outperforms the baseline across all strata, the gain is
smallest under high occlusion: in the hardest stratum of each factor, the
improvement is only $0.5$~mm for occlusion, compared with $1.1$~mm for both
person scale and inter-frame motion. This suggests that severe occlusion
limits the additional benefit that can be obtained from the proposed method.

% Tab.~\ref{tab:supp_failure} reveals three clear trends. First, weaker visual
% observations lead to larger reconstruction errors. For person scale, UT-HMR's
% MPJPE increases from $36.33$~mm for the largest quartile to $51.01$~mm for the
% smallest. Occlusion has an even stronger effect, with the error increasing from
% $37.48$~mm for non-overlapping persons to $49.93$~mm for overlapping persons
% and reaching $61.88$~mm under heavy overlap. Second, larger inter-frame motion
% also degrades performance, with MPJPE increasing from $35.53$~mm for the
% slowest quartile to $48.91$~mm for the fastest. These results identify
% occlusion and weak visual evidence as the most challenging conditions for
% reconstruction.

% Despite these challenges, UT-HMR consistently outperforms SAT-HMR across all
% strata. The improvement is larger for moderate motion ($1.32$~mm) but decreases
% to $1.05$~mm for the fastest-moving subjects. Likewise, the gain drops to only
% $0.47$~mm under heavy occlusion, suggesting that a short temporal window
% provides limited additional evidence when neighboring frames are poorly
% aligned or strongly occluded.

\subsection{Representative Failure Cases}

Fig.~\ref{fig:supp_failure} shows two representative failure cases that illustrate the limitations
identified above. The top example is a small, heavily motion-blurred player in ice hockey, where
the input provides limited visual evidence for reliable pose estimation. The bottom example is
from 3DPW, where severe occlusion is combined with an uncommon pose. In both cases, UT-HMR makes substantial pose errors in specific body
parts, highlighting that temporal fusion and uncertainty modeling cannot fully resolve ambiguity
when the available visual evidence is severely degraded or atypical.

\begin{figure}[t]
    \centering
    \includegraphics[width=\linewidth]{figs/supp_failure.png}
    \caption{Representative failure cases of UT-HMR. Top: a small, heavily
    motion-blurred player in ice hockey. Bottom: a person entering a car
    in 3DPW, combining severe occlusion with an uncommon pose.}
    \label{fig:supp_failure}
\end{figure}

\section{Additional Implementation Details}
\label{sec:supp_impl}

\subsection{Architecture and Temporal Early Fusion}
UT-HMR is built on SAT-HMR with a DINOv2 ViT-B/14 backbone, retaining its input
resolution and scale-adaptive token scheme. For a $T{=}3$ frame clip
($t{-}1$, $t$, $t{+}1$), frames are independently tokenized by the shared
encoder, and the three token sequences are concatenated into a single memory.
The decoder then uses the center-frame person queries to cross-attend to
this fused memory, injecting temporal evidence before pose decoding while
keeping the query and prediction heads unchanged. Only the center frame is
supervised during training.

\subsection{Uncertainty Head}

For each query, the final decoder feature is fed into two MLP heads to predict
the low-rank-plus-diagonal covariance parameters. One head outputs
$\mathbf{L}\in\mathbb{R}^{144\times r}$, with weights initialized from a
zero-mean Gaussian with standard deviation $10^{-3}$ and zero bias, while the
other predicts $\log\mathbf{d}\in\mathbb{R}^{144}$, initialized with zero
weights and bias $-2$. Thus, the initial diagonal variance is $e^{-2}$, while
the small random initialization of $\mathbf{L}$ avoids the zero-gradient state
at $\mathbf{L}=\mathbf{0}$, where the NLL gradient with respect to $\mathbf{L}$
vanishes.

For numerical stability, we clamp the entries of $\mathbf{L}$ to $[-3,3]$ and
$\log\mathbf{d}$ to $[-4,2]$, and add a diagonal floor of $10^{-3}$ when
evaluating the covariance. These constraints help prevent extreme covariance
values and numerical instability in the Cholesky decomposition used for the
Gaussian NLL. The NLL is computed on the residual between the ground-truth and
predicted 6D pose, with the predicted mean detached. Hence, the uncertainty
branch learns to model the residual uncertainty of the deterministic pose
estimate without directly updating the pose mean through the distribution loss.

\section{Supplementary Video}
\label{sec:supp_video}

We provide two types of visualization videos for all three benchmarks:

\begin{itemize}
    \item The first set of videos
(\texttt{icehockey\_compare.mp4}, \texttt{soccer\_compare.mp4}, and
\texttt{3dpw\_compare.mp4}) compares, from left to right, the input, SAT-HMR (baseline),
UT-HMR (ours), and ground-truth meshes. These videos highlight the improvements of UT-HMR
over the baseline on the ice hockey and soccer benchmarks, while showing similar results
on 3DPW. For clearer visualization, the ice hockey and soccer videos crop the target
person region and render only the target player; these crops are used solely for
visualization and are not provided as inputs during inference.

\item The second set (\texttt{icehockey.mp4}, \texttt{soccer.mp4}, and
\texttt{3dpw.mp4}) shows the original full-frame input on the left and the
corresponding UT-HMR reconstruction on the right. These videos provide a direct
visualization of the full-image one-stage inference and the temporal behavior of
UT-HMR across different datasets.

\end{itemize}
{
    \small
    \bibliographystyle{ieeenat_fullname}
    % \bibliography{main}
}